\documentclass[10pt,twocolumn,letterpaper]{article}

\usepackage{cvpr}
\usepackage{times}
\usepackage{epsfig}
\usepackage{graphicx}
\usepackage{amsmath}
\usepackage{amssymb}

\usepackage{subcaption}
\usepackage{booktabs}
\usepackage[british,american]{babel}
\usepackage{amsfonts}
\usepackage{xcolor}
\usepackage{enumitem}
\usepackage[pagebackref=false,breaklinks=true,letterpaper=true,colorlinks,bookmarks=false]{hyperref}
\usepackage{textpos}
\let\oldparagraph\paragraph
\renewcommand{\paragraph}[1]{\vspace{-0.27cm} \oldparagraph{#1}}
\newcommand{\figref}[1]{\mbox{Fig.~\ref{#1}}}
\newcommand{\tblref}[1]{\mbox{Table~\ref{#1}}}
\newcommand{\secref}[1]{\mbox{Sec.~\ref{#1}}}
\renewcommand{\eqref}[1]{\mbox{Eq.~\ref{#1}}}
\newcommand*{\email}[1]{\tt\small{#1}}
\newcommand*{\affmark}[1][*]{\textsuperscript{#1}}

\cvprfinalcopy 

\def\cvprPaperID{}

\ifcvprfinal\fi

\begin{document}

\title{From Paris to Berlin: Discovering Fashion Style Influences Around the World}

\author{
\setlength{\tabcolsep}{10pt}
\begin{tabular}{cc}
Ziad Al-Halah\affmark[1] & Kristen Grauman\affmark[1,2]\\
\email{ziadlhlh@gmail.com} & \email{grauman@cs.utexas.edu}\\
\affmark[1]The University of Texas at Austin & \affmark[2]Facebook AI Research
\end{tabular}
}

\maketitle

\begin{abstract}
The evolution of clothing styles and their migration across the world is intriguing, yet difficult to describe quantitatively.
We propose to discover and quantify fashion influences from everyday images of people wearing clothes.
We introduce an approach that detects which cities influence which other cities in terms of propagating their styles.
We then leverage the discovered influence patterns to inform a forecasting model that predicts the popularity of any given style at any given city into the future.
Demonstrating our idea with GeoStyle---a large-scale dataset of 7.7M images covering 44 major world cities, we present the discovered influence relationships, revealing how cities exert and receive fashion influence for an array of 50 observed visual styles.
Furthermore, the proposed forecasting model achieves state-of-the-art results for a challenging style forecasting task, showing the advantage of grounding visual style evolution both spatially and temporally. Project page: \url{https://www.cs.utexas.edu/~ziad/fashion_influence.html}.
\end{abstract}
 
\begin{textblock*}{\textwidth}(0cm,-15cm)
\centering\small
IEEE Conference on Computer Vision and Pattern Recognition (CVPR), 2020.
\end{textblock*}

\section{Introduction}

\emph{``The influence of Paris, for instance, is now minimal. Yet a lot is written about Paris fashion."}---Geoffrey Beene
\vspace{0.1in}

The clothes people wear are a function of personal factors like comfort, taste, and occasion---but also wider and more subtle influences from the world around them, like changing social norms, art, the political climate, celebrities and style icons, the weather, and the mood of the city in which they live.
Fashion itself is an evolving phenomenon because of these changing influences.
What gets worn continues to change, in ways fast, slow, and sometimes cyclical.

Pinpointing the influences in fashion, however, is non-trivial.
To what extent did the runway styles in Paris last year affect what U.S. consumers wore this year?  How much did the designs by J. Crew influence those created six months later by Everlane, and vice versa?  How long does it take for certain trends favored in New York City to migrate to Austin, if they do at all?  And how did the infamous cerulean sweater worn by the protagonist in the movie \emph{The Devil Wears Prada} make its way into her closet?\footnote{\color{darkgray}{The Devil Wears Prada: Cerulean~\href{https://bit.ly/3dBAQ5W}{https://bit.ly/3dBAQ5W}}}

To quantitatively answer such questions would be valuable to both social science and business forecasts, yet it remains challenging. 
 Clothing sales records or social media ``likes" offer some signal about how tastes are shifting, but they are indirect and do not reveal the sources of influence.

\begin{figure}[t]
\centering
    \includegraphics[width=0.98\linewidth]{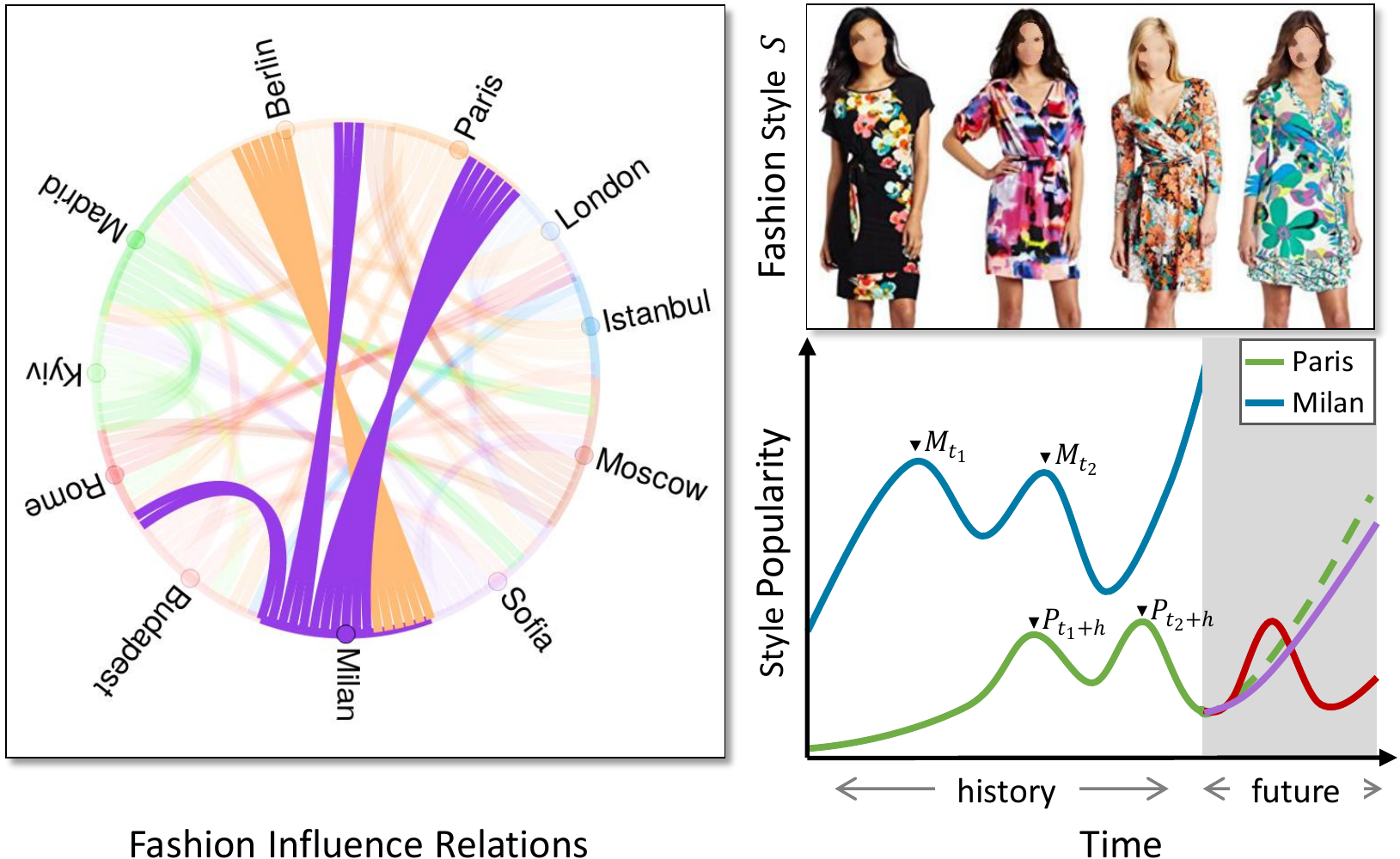}
\caption{Styles propagate according to certain patterns of influence around the world.  For example, the trajectory of a given style's popularity in Milan may foreshadow its trajectory in Paris some months later.  Our idea is to discover style influence relations worldwide (left) and leverage them to accurately forecast future trends per location (right).  Whereas forecasting without regard to geographic influence can falter in the presence of complex trends (red curve), using discovered influence information (\eg, \emph{Milan} influences \emph{Paris} in fashion style $S$) yields better forecasts (purple curve).
}
\label{fig:intro}
\end{figure}

We contend that images are exactly the right data to answer such questions.
Unlike vendors' purchase data, 
other non-visual metadata, or hype from haute couture designers,
everyday photos of what people are wearing in their daily life provide a unfiltered glimpse of current clothing styles ``on the ground".
Our idea is to discover fashion influence patterns in community photo collections (\eg, Instagram, Flickr), and leverage those influence patterns to forecast future style trends conditioned on the place in the world.
While fashion influences exist along several axes, we focus on worldwide geography to capture spatio-temporal influences.
Specifically, we aim to discover \emph{which cities influence which other cities} in terms of propagating their clothing styles, and \emph{with what time delay}.

To this end, we introduce an approach to discover geographical style influences from photos.
First, we extract a vocabulary of visual styles from unlabeled, geolocated and timestamped photos of people.
Each style is a mixture of detected visual attributes.
For example, one style may capture  short floral dresses in bright colors (Fig.~\ref{fig:intro}) while another style captures preppy collared shirts.
Next, we record the past trajectories of the popularity of each style, meaning the frequency with which it is seen in the photos over time.
Then, we identify two key properties of an influencer---time precedence and novelty---and use a statistical measure that captures these properties to calculate the degree of influence between cities.
Next, we introduce a neural forecasting model that exploits the influence relationships discovered from photos to better anticipate future popular styles in any given location.
Finally, we propose a novel coherence loss to train our model to reconcile the local predictions with the global trend of a style for consistent forecasts.
We demonstrate our approach on the large-scale GeoStyle dataset~\cite{mall2019geostyle} comprised of everyday photos of people with a wide coverage of geographic locations.  

Our results shed light on the spatio-temporal migration of fashion trends across the world---revealing which cities are exerting and receiving more influence on others, which most affect global trends, which contribute to the prominence of a given style, and how a city's degree of influence has itself changed over time.
Our findings hint at how computer vision can help democratize our understanding of fashion influence, sometimes challenging common perceptions about what parts of the world are driving fashion (consistent with designer Geoffrey Beene's quote above).  

In addition, we demonstrate that by incorporating influence, the proposed forecasting model yields state-of-the-art accuracy for predicting the future popularity of styles.
Unlike prior work that learns trends with a monolithic worldwide model~\cite{Al-Halah2017} or independent per city models~\cite{mall2019geostyle}, our geo-spatially grounded predictions catch the temporal dependencies between when different cities will see a style climb or dip, producing more accurate forecasts (see Fig.~\ref{fig:intro}).

\section{Related Work}

Visual fashion analysis, with its challenging vision problems and  direct impact on our social and financial life, presents an attractive domain for vision research.
In recent years, many aspects of fashion have been addressed in the computer vision literature, ranging from learning fashion attributes~\cite{berg2010automatic,bossard2012apparel,chen2012describing,Chen2015,liu2016}, landmark detection~\cite{Wang_2018_CVPR,Yu_2019_CVPR}, cross-domain fashion retrieval~\cite{huang2015cross,hadi2015buy,Zhao_2017_CVPR,Kuang_2019_ICCV}, body shape and size based fashion suggestions~\cite{misra2018decomposing,hidayati2018dress,kimberly-cvpr2020}, virtual try-on~\cite{wang2018toward,Dong_2019_ICCV}, clothing recommendation~\cite{liu2012hi,mcauley2015image,Yu_2019_ICCV,hsiao2019fashion++}, inferring social cues from people's clothes~\cite{song2011predicting,murillo2012urban,kwak2013bikers}, outfit compatibility~\cite{hsiao2018creating,Han_2019_ICCV}, visual brand analysis~\cite{kim2013discovering,hadi2018brand}, and discovering fashion styles~\cite{kiapour2014hipster,veit2015learning,Al-Halah2017,hsiao2017latent-look}.
Our work opens a new avenue for visual fashion understanding: 
modeling influence relations in fashion directly from images. 

\vspace*{-0.1in}
\paragraph{Statistics of styles} 
Analyzing styles' popularity in the past gives a window on people's preferences in fashion.  
Prior work considers how the frequency of attributes (\eg, floral, neon) changed over time~\cite{Vittayakorn2015}, and how trends in (non-visual) clothing meta-data changed for the two cities Manila and Los Angeles~\cite{Simo-Serra2015}.
Qualitative studies suggest how collaborative filtering recommendation models can account for past temporal changes of fashion~\cite{He2016} or what cities exhibit strong style similarities~\cite{kataoka2019ten}.  
However, all this prior work analyzes style popularity in an ``after the fact'' manner, and looks only qualitatively at past changes in style trends.
We propose to go beyond this historical perspective to forecast future changes in styles' popularity along with supporting quantitative evaluation.

\vspace*{-0.1in}
\paragraph{Trend forecasting}

Only limited prior work explores forecasting visual styles into the future~\cite{Al-Halah2017,mall2019geostyle}.
The FashionForward model~\cite{Al-Halah2017} uses fashion styles learned from Amazon product images to train an exponential smoothing model for forecasting, treating the products' transaction history (purchases)  as a proxy for style popularity.
Similarly, the GeoStyle project~\cite{mall2019geostyle} uses a seasonal forecasting model to predict changes in style trends per city and highlight unusual events. 
Both prior models assume that style trends in different cities are independent from one another and can be modeled monolithically~\cite{Al-Halah2017} or in isolation~\cite{mall2019geostyle}.
In contrast, we introduce a novel model that accounts for influence patterns discovered across different cities. 
Our concept of fashion influence discovery is new, and our resulting forecasting model outperforms state-of-the-art.

\vspace*{-0.1in}
\paragraph{Influence modeling}
To our knowledge, no previous work tackles influence modeling in the visual fashion domain.
The closest study 
looks at the correlation among attributes popular in New York fashion shows and those attributes seen in street photos, as a surrogate for fashion shows' impact on people's clothing; however, no influence or forecasting model is developed~\cite{Chen2015}.
Outside the fashion domain, models for influence are developed for connecting text in news articles~\cite{shahaf2010connecting}, linking video subshots for summarization~\cite{lu2013story}, or analyzing intellectual links between major AI conferences from their papers~\cite{chen2018modeling}.
Our method is the first to model influence in visual fashion trends.
We propose an influence model that is grounded by forecasting accuracy.
Through our evaluation, we show that our model discovers interesting influence patterns in fashion that go beyond simple correlations, and we analyze influence 
on multiple axes to discover locally and globally influential players.

\section{Visual Style Influence Model}\label{sec:approach}

We propose an approach to model influence in the visual fashion domain.
Starting with images of fashion garments, 1) we learn a visual style model that captures the fine-grained properties common among the garments;
then 2) we construct style popularity trajectories by leveraging images' temporal and spatial meta information;
3) we model the influence relations between different locations (cities) for a given visual style;
Finally, 4) we introduce a forecasting model that utilizes the learned influence relations together with 
a coherence loss for consistent and accurate predictions of future changes in style popularity.

\subsection{Visual Fashion Styles}\label{sec:app_style}
Our model captures the fashion influence among different locations in the world.
We begin by discovering a set of visual fashion \emph{styles} from images of people's garments in their everyday life.  As discussed above, such photos offer an unfiltered glimpse of what people are wearing around the world.  We use the 7.7M-image GeoStyle dataset~\cite{mall2019geostyle} from Instagram and Flickr as source data. 

Let $X=\{x_i\}^N$ be a set of clothing images.  We first learn a semantic representation that captures the elementary fashion attributes like colors (\eg cyan, green), patterns (\eg stripes, dots), shape (\eg v\-neck, sleeveless) and garment type (\eg shirt, sunglasses).
Given a fashion attribute model $f_a(\cdot)$ (\eg, a CNN) trained on a set of disjoint labeled images, we can then represent each image in $X$ with $\mathbf{a}_i=f_a(x_i)$, where $\mathbf{a}\in\mathbb{R}^M$ is a vector of $M$ visual attribute probabilities.

\begin{figure}[t]
\centering
    \includegraphics[width=1.\linewidth]{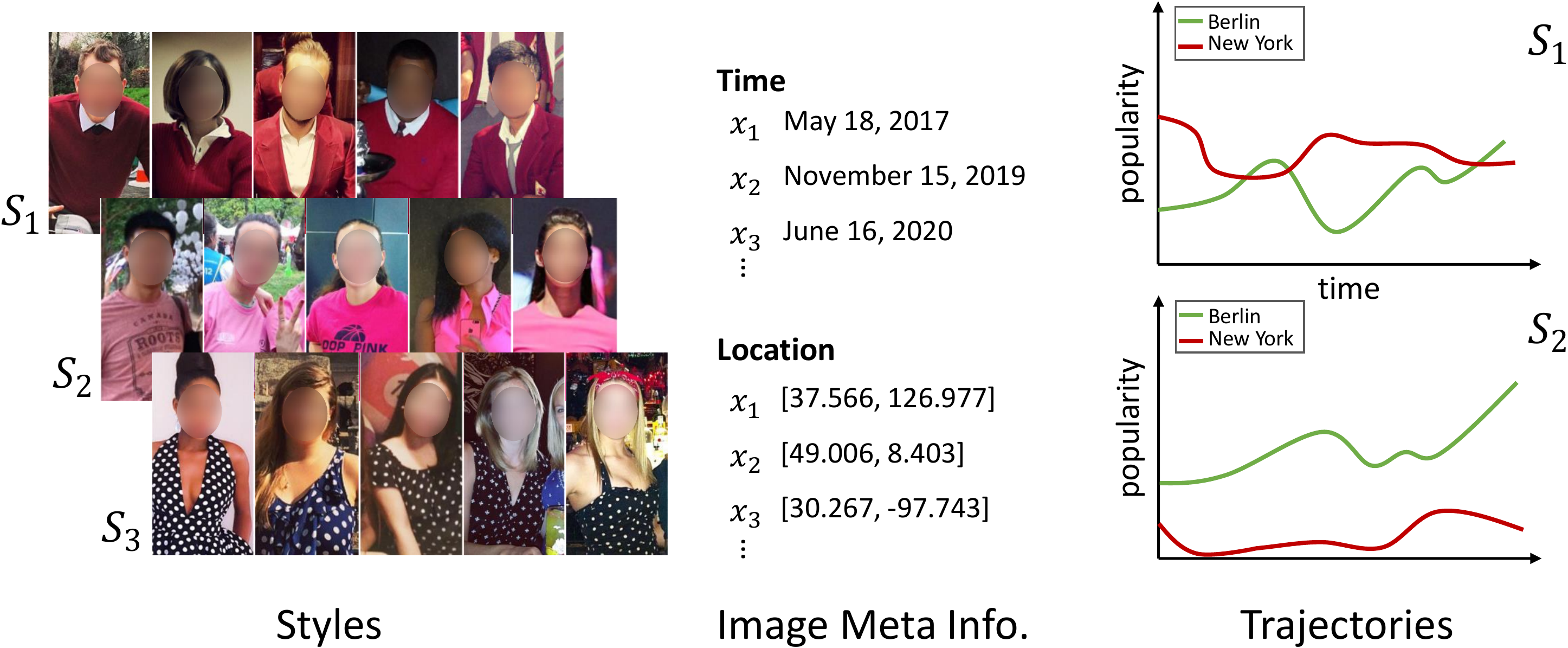}
\caption{Style trajectories.
First, we learn a set of fashion styles from everyday images (left).  Then based on images' timestamps and geolocations (center) we measure the popularity of a style at a given place (\eg city) and a time period (\eg week) to build up the style popularity trajectory (right).
}
\label{fig:styles_traj}
\end{figure}

Next, we learn a set of fashion styles $S=\{S_k\}^K$ that capture distinctive attribute combinations using a Gaussian mixture model (GMM) of $K$ components.  
\figref{fig:styles_traj} shows a set of three fashion styles discoverd by this style model from Instagram images.
Hence, given an image of a new garment $x_i$, the style model $f_s(\cdot)$ can predict the probabilities of that garment to be from each of the learned styles $\mathbf{s}_i=f_s(\mathbf{a}_i)$.

\paragraph{Style trajectories}\label{sec:app_traj}

We measure the popularity of a fashion style in a certain location through the frequency of the style in the photos of the people in that location.
Specifically, given the timestamps and geolocations of the photos, we first quantize them into a meaningful temporal resolution (\eg weeks, months) and locations (\eg cities).
Then, we construct a temporal trajectory $y^{ij}$ for each pair of style and location $(S^i, C^j)$:
\begin{equation}
    y^{ij}_{t} = \frac{1}{|C_{t}^{j}|} \sum_{x_k \in C_{t}^{j}} p(S^i|x_k),
\end{equation}
where $C_t^j$ is the set of images from location $C^j$ in the time window $t$, $p(S^i|x_k)$ is the probability of style $S^i$ given image $x_k$ based on our style model $f_s(\cdot)$, and $y^{ij}_t$ is the popularity of style $S^i$ in location $C^j$ during time $t$.
Finally, by getting all values for $t={1,\dots, T}$ we construct the temporal trajectory $y^{ij}$.  See \figref{fig:styles_traj}.

The GeoStyle dataset, like any Internet photo dataset, it has certain biases in terms of the demographics of the people who have uploaded photos and the locations---as discussed by the dataset creators~\cite{matzen2017streetstyle}.
These biases may affect the type of styles considered and their measured popularity.
For example, younger generations are more likely to upload photos to Instagram and from places with easy access to high Internet bandwidth.
Nonetheless, the dataset is the largest public fashion dataset with the most temporal and geographic coverage, providing a unique glimpse on people's fashion preferences around the globe.

Next, we describe our influence model that analyzes these trajectories to discover the influence patterns among the various locations.

 \subsection{Influence Modeling}\label{sec:app_influence}

We propose to ground fashion influence through style popularity forecasting.
This enables us to quantitatively evaluate influence using learned computational models based on real world data.

We say city $C^i$ influences city $C^j$ in a given fashion style $S^n$ if our ability to accurately forecast the popularity of $S^n$ in $C^j$ significantly improves when taking into consideration the past popularity trend of $S^n$ in $C^i$, in addition to its past popularity trend in $C^j$.
In other words, past observations in $y^{ni}_{1\dots t}$ provide us with new insight on the future changes in $y^{nj}_{t+1\dots t+h}$ that are not available in $y^{nj}_{1\dots t}$. 

We identify two main properties of the influencer $C^i$:
1) time precedence, that is the influencer city's changes happen before the observed impact on the influenced city and 2) novelty, that is the influencer city has novel past information not observed in the history of the influenced city.

A naive approach to capture such relations is to use a multivariate model to learn to predict $y^{nj}_t$ by feeding it all available information from the other cities.
However, this approach does not satisfy the second property for an influencer since it does not constrain the influencer to have novel information that is not present in the influenced.
Instead, our fashion influence relations can be captured using the Granger causality test~\cite{granger1969}.
The test determines that a time series $y^1$ Granger causes a time series $y^2$ if, while taking into account the past values of $y^2_{1\dots t}$, the past values of $y^1_{1\dots t}$ still have statistically significant impact on predicting the next value of $y^2_{t+1}$.
The test proceeds by modeling $y^2$ with an autoregressor of order $d$, \ie:
\begin{equation}
    y^2_t = \phi_0 + \sum_{k=1}^d \phi_k y^2_{t-k} + \sigma_t,
\end{equation}
where $\sigma_t$ is an error term and $\phi_k$ contains the regression coefficients.
Then the autoregressor of $y^2$ is extended with lagged values of $y^1$ such that:
\begin{equation}
    y^2_t = \phi_0 + \sum_{k=1}^d \phi_k y^2_{t-k} + \sum_{l=m}^q \psi_l y^1_{t-l} + \sigma_t.
\end{equation}
If the extended lags from $y^1$ do add significant explanatory power to $y^2_t$, \ie the forecast accuracy of $y^2$ is significantly better ($p<0.05$) according to a regression metric (mean squared error), then $y^1$ Granger causes $y^2$.

We estimate the influence relations across all cities' trajectories for each fashion style $S^i$.  (In experiments we consider $50$ such styles, and consider lags ranging from $1$ to $8$ temporal steps.)
In this way, we establish the influence relations among cities and at which lag this influence occurs.

\begin{figure}[t]
\centering
    \includegraphics[width=1.\linewidth]{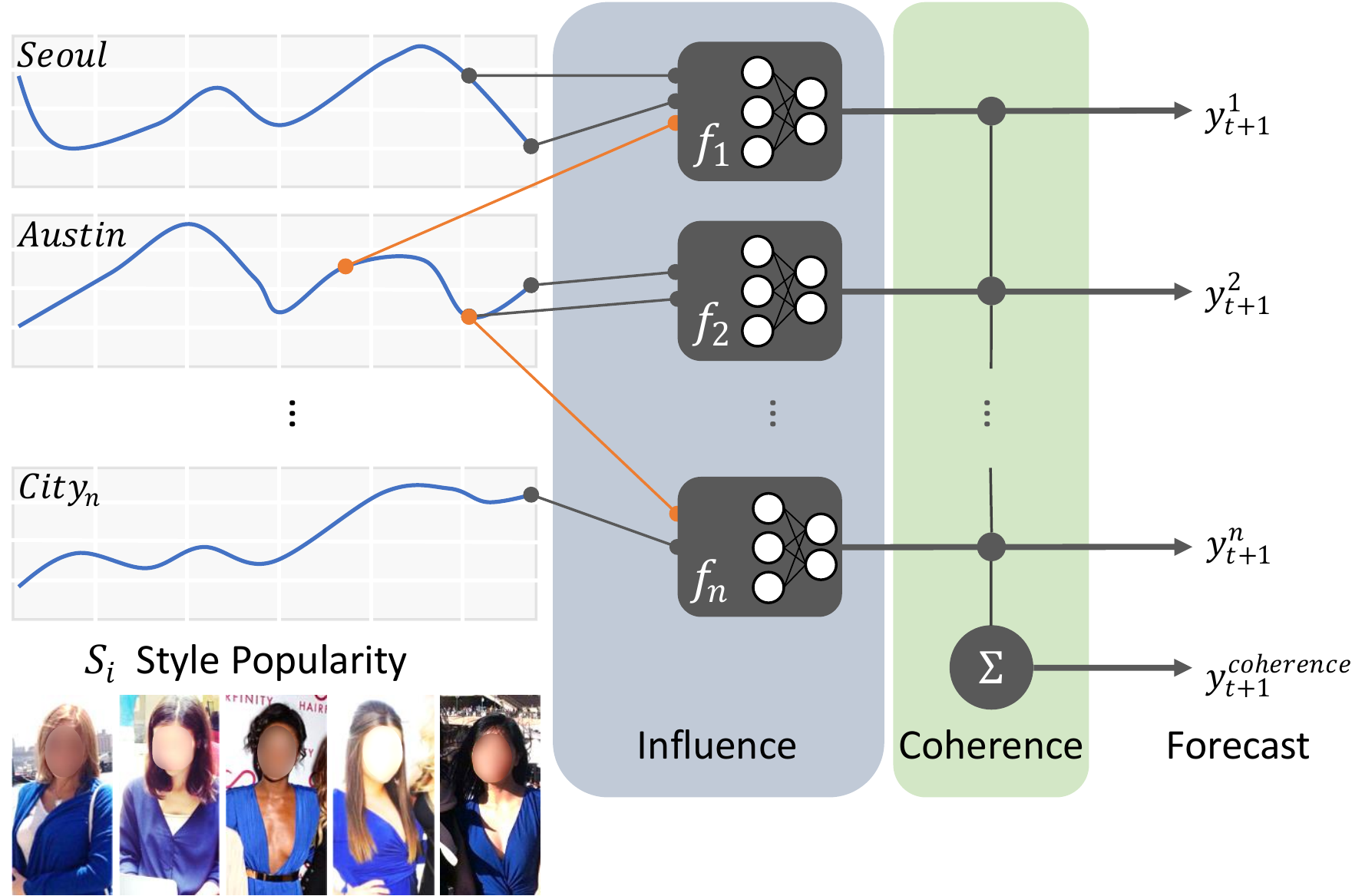}
\caption{Influence coherent forecaster.
Our model captures influence relations between cities for a given fashion style (orange connections) and 
uses them to predict future changes in the style popularity in each location.
Additionally, our model regularizes the forecasts to be coherent with the global trend of the style observed across all cities.}
\label{fig:model}
\vspace{-0.2cm}
\end{figure}
 \subsection{Coherent Style Forecaster}\label{sec:app_forecast}

After we estimate the influence relations across the cities, we build a forecaster for each trajectory $y^{ij}$ such that:
\begin{equation}
    \tilde{{y}}^{ij}_{t+1} = f(L(y^{ij}_t),I(y^{ij}_t) | \theta),
\end{equation}
where $I(y^{ij}_t)$ is the set of lags from the influencer of $y^{ij}$ relative to time step $t$ as determined in the previous section, and $L(y^{ij}_t)$ are the lags from $y^{ij}$'s own style popularity trajectory.
We model $f(\cdot)$ using a multilayer perceptron (MLP) and estimate the parameters $\theta$ by minimizing the mean squared error loss:
\begin{equation}\label{eq:loss_forecast}
    \mathcal{L}_{forecast} = \sum_t (y^{ij}_{t+1} - f(L(y^{ij}_t),I(y^{ij}_t) | \theta))^2,
\end{equation}
where $y^{ij}_{t+1}$ is the ground truth value of $y^{ij}$ at time $t+1$.

\paragraph{Coherence loss}\label{sec:app_coherence}

Our forecast model in its previous form does not impose any constraints on the forecasted values in relation to each other.
However, while we are forecasting the style popularity at each individual location given the influence from the others, the forecasted popularities ($y^{i1}_{t}, y^{i2}_{t} \dots y^{in}_{t}$) are still for a common fashion style $S^i$ that by itself exhibits a worldwide trend across all locations.

We propose to reconcile the base forecasts produced at each location through a \emph{coherence loss} that captures the global trend.
For all forecasts $\hat{y}^{ij}_{t+1}$ for a fashion style $S^i$ and across all cities $C^j\in C$, we constrain the distribution mean of the predicted values to match the distribution mean of the ground truth values:
\begin{equation}\label{eq:loss_coherence}
    \mathcal{L}_{coherence} = \frac{1}{|C|} (\sum_k y^{ik}_{t+1} - \sum_k \hat{y}^{ik}_{t+1}).
\end{equation}

The coherence loss, in addition to capturing the global trend of $S^i$, helps in combating noise at the trajectory level of each city through regularizing the mean distribution of all forecasts.
The final model is trained with the combined forecast and coherence losses:
\begin{equation}\label{eq:loss}
    \mathcal{L} = \mathcal{L}_{forecast} + \lambda \mathcal{L}_{coherence}.
\end{equation}
\figref{fig:model} illustrates our model.
For a style $S_i$, we model its popularity trajectory at each location (\eg, Seoul) with a neural network of $2$ layers and sigmoid non-linearity.
The input of the network is defined by the lags from its own trajectory (shown in black) and any other influential lags from other cities discovered by the previous step (\secref{sec:app_influence}) which are shown in orange.
Furthermore, the output of all local forecasters is further regularized to be coherent with the overall observed trend of $S_i$ using our coherence loss.

\section{Evaluation}

In the following experiments, we demonstrate our model's ability to forecast styles' popularity changes around the globe using the discovered influence relations.
Furthermore, we analyze the influence patterns revealed by our model between major cities, how they influence global trends, and their dynamic influence trends through time.

\paragraph{Dataset}
We evaluate our approach on the GeoStyle dataset~\cite{mall2019geostyle} which is based on Instagram and Flickr photos showing people from $44$ major cities from around the world.
In total, the dataset has $7.7$ million images that span a time period from July 2013 until May 2016.
The dataset is used for research purposes only.

\paragraph{Styles and popularity}
We use attribute predictions from~\cite{mall2019geostyle} to represent each photo with $46$ fashion attributes (\eg colors, patterns and garment types).
Based on these, we learn $50$ fashion styles using a Gaussian mixture model.
Then, for each style we infer its popularity trajectories in each city using a temporal resolution of weeks (cf.~\secref{sec:app_style}).
While we focus on fashion styles in this work (each of which aggregates an array of commonly co-occurring attributes), we find similar results when considering individual visual attributes as the fashion concept of interest.

\paragraph{Implementation details}
We set $\lambda=1$ for the coherence loss weight (see \eqref{eq:loss}) and optimize our neural influence model using Adam for stochastic gradient descent with a learning rate of $10^{-2}$ and $l_2$ weight regularization of $10^{-8}$.
We select the best model based on the performance on a disjoint validation split using early stopping.

\subsection{Influence-based Forecasting}

We evaluate how well our model produces accurate forecasts by leveraging the influence pattern, and compare it to several baselines and existing methods~\cite{Al-Halah2017,mall2019geostyle} that model trajectories in isolation.
We adopt a \emph{long-term forecasting} setup where we use the last $26$ points from each style trajectory for testing, and the rest for model training.
We arrange the baselines into three main groups:

\vspace{0.1cm}
\noindent\textbf{Naive models}: these models rely on basic statistical properties of the trajectory to produce a forecast.
We consider five variants of these baselines similar to those from \cite{Al-Halah2017}; see appendix for a detailed description.

\noindent\textbf{Per-City models}:
These fit a separate parametric model trained on the history ``lags'' of each of the trajectories~\cite{box2015time}. 
\begin{itemize}[leftmargin=*,itemsep=2.5pt,labelsep=2pt,label=\textendash]
\vspace{-0.21cm}
\setlength{\parskip}{0pt}
\setlength{\parsep}{0pt}
    \item  FashionForward (EXP)~\cite{Al-Halah2017}: an exponential decay model which forecasts based on a learned weighted average of the historical values.
    \item AR: it fits a standard autoregression model.
    \item ARIMA: the standard autoregressive integrated moving average model.
    \item  GeoModel~\cite{mall2019geostyle}: a parametric seasonal forecaster.
\end{itemize}
To our knowledge, the two existing methods~\cite{Al-Halah2017,mall2019geostyle} represent the only prior vision approaches for style forecasting.
Further, unlike our approach, all of the per-city models consider the popularity trajectories of the styles in isolation, \ie, they do not take into consideration possible interactions among the cities. 

\noindent\textbf{All-Cities models}:
These fit a parametric model trained on the trajectories of a style across all cities. Such models assume a full and simultaneous interaction between all cities.
We consider the VAR model~\cite{box2015time} to represent this group.

\begin{table}[t]
\setlength{\tabcolsep}{8pt}
\center
\scalebox{0.78}{
\begin{tabular}{l r r r r}
\toprule
    & \multicolumn{2}{c}{Seasonal} & \multicolumn{2}{c}{Deseasonalized}\\
    Model               &  MAE     &   MAPE &  MAE     &   MAPE \\
\midrule
    \multicolumn{5}{l}{\textbf{Naive}}  \\
    Gaussian            &   0.1301   &   33.23               &   0.1222  &   26.08    \\
    Seasonal            &   0.0925   &   22.64               &   0.1500 &   33.39    \\
    Mean                &   0.0908   &   23.57               &   0.0847  &   18.97    \\
    Last                &   0.0893   &   22.20               &   0.1053 &   23.08    \\
    Drift               &   0.0956   &   23.65               &   0.1163 &   25.32   \\
\midrule
    \multicolumn{5}{l}{\textbf{Per City}} \\
    FashionForward (EXP)~\cite{Al-Halah2017}
                        &   0.0779   &   19.76               &   0.0848   &   18.94    \\
    AR                  &   0.0846   &   21.88               &   0.0846   &   18.95    \\ 
    ARIMA               &   0.0919   &   23.70               &   0.1033  &  22.70    \\ 
    GeoModel~\cite{mall2019geostyle}
                        &   0.0715   &   17.86               &   0.0916   &   20.31    \\
\midrule
    \multicolumn{5}{l}{\textbf{All Cities}}\\
    VAR                 &   0.0771   &   19.25               &   0.0929   &   20.41   \\ 
\midrule
    \multicolumn{5}{l}{\textbf{Ours -- Influence-based}}   \\
    Full                &   \textbf{0.0699}  &   \textbf{17.38}   &   \textbf{0.0824}    &   \textbf{18.29}    \\ 
    w/o Influence       &    0.0708   &   17.70              &    0.0859   &   19.24    \\ 
    w/o Influence \& Coherence  &   0.0858  &   20.95        &    0.0942   &     20.62  \\ 
\bottomrule
\end{tabular}
}
\vspace*{-0.1in}
\caption{Forecast errors on the GeoStyle dataset~\cite{mall2019geostyle} for seasonal and deseasonalized fashion style trajectories. } 
\label{tbl:forecast_geostyle_ext}
\end{table}
 
We compare all models using the forecast error captured by the mean absolute error (MAE), which measures the absolute difference between the forecasted and ground truth values, and the mean absolute percentage error (MAPE), which measures the forecast error scaled by the ground truth values.  
Additionally, to quantify the impact of possible seasonal yearly trends in fashion styles, we also consider forecasting the deseasonalized style trajectories (\ie we subtract the yearly seasonal lag  from the trajectories).  
The deseasonalized test is interesting because it requires methods to capture the subtle visual trends not simply associated with the location's weather and annual events.

\begin{figure*}[t]
\centering
\begin{subfigure}{.32\textwidth}
    \centering
    \includegraphics[trim={3cm 4cm 3.5cm 3.5cm} ,clip,width=0.98\linewidth]{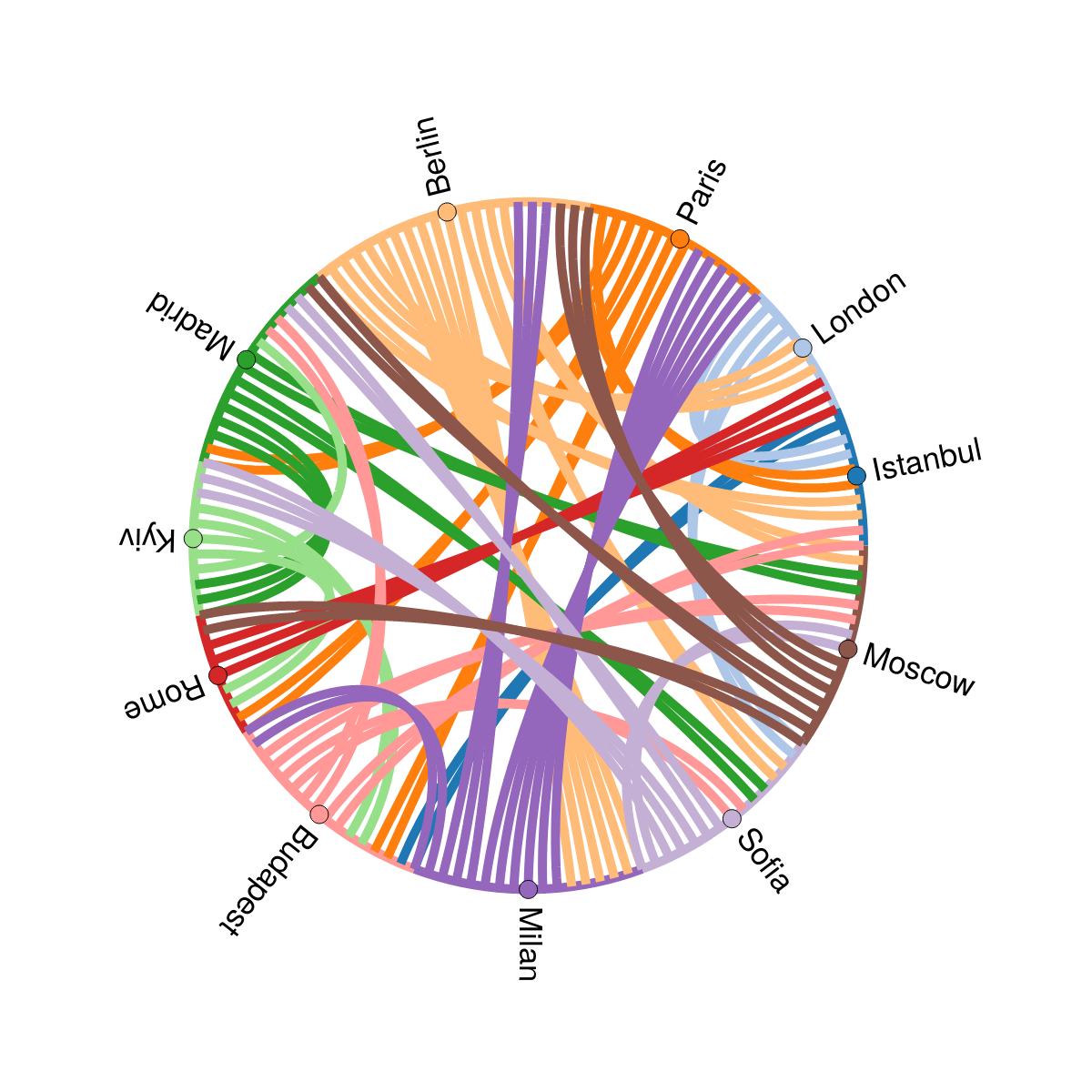}
    \caption{European Cities}\label{fig:infl_cities_style_a}
\end{subfigure}
\begin{subfigure}{.32\textwidth}
    \centering
    \includegraphics[trim={3cm 4cm 3.5cm 3.5cm} ,clip,width=0.98\linewidth]{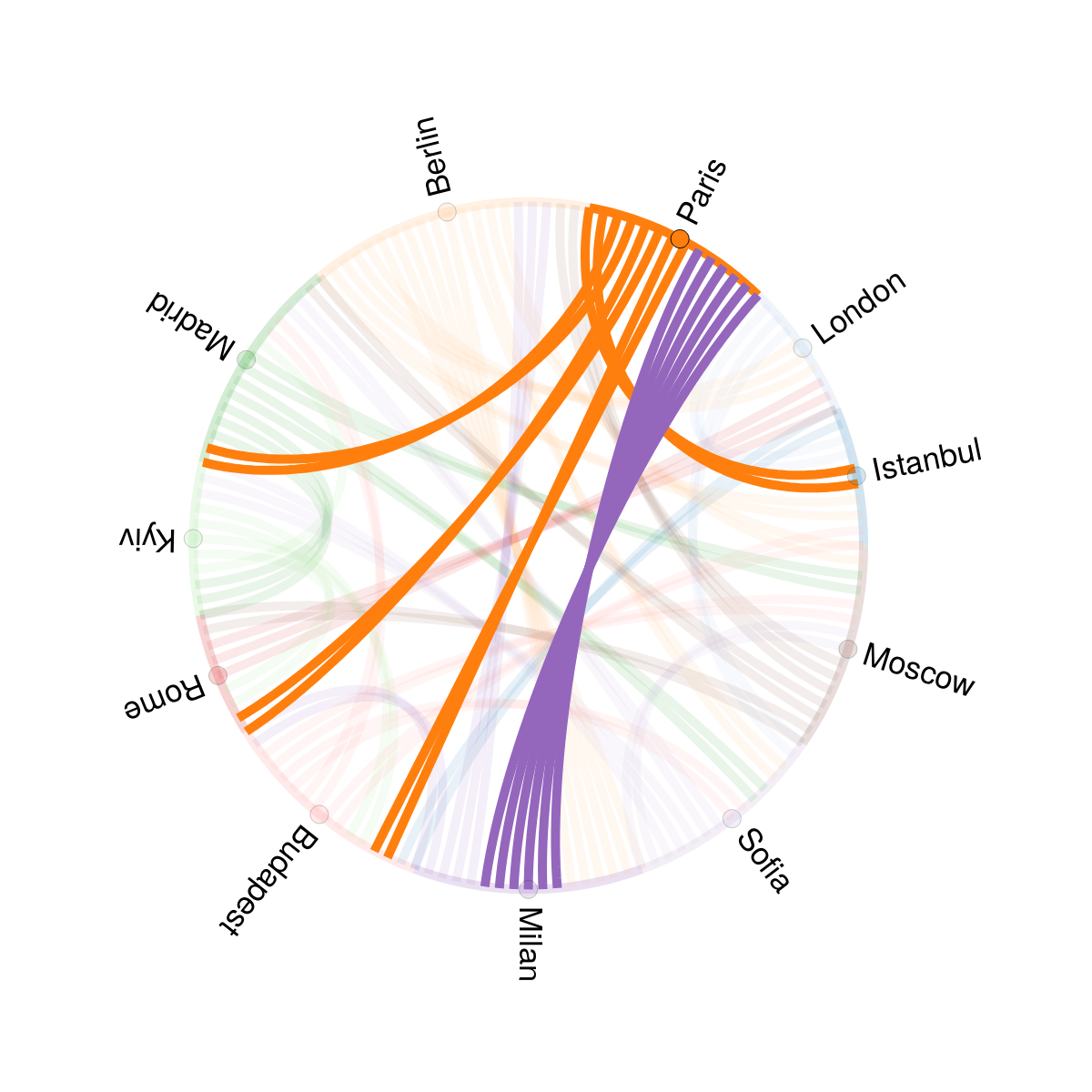}
    \caption{Paris}\label{fig:infl_cities_style_b}
\end{subfigure}
\begin{subfigure}{.32\textwidth}
    \centering
    \includegraphics[trim={3cm 4cm 3.5cm 3.5cm} ,clip,width=0.98\linewidth]{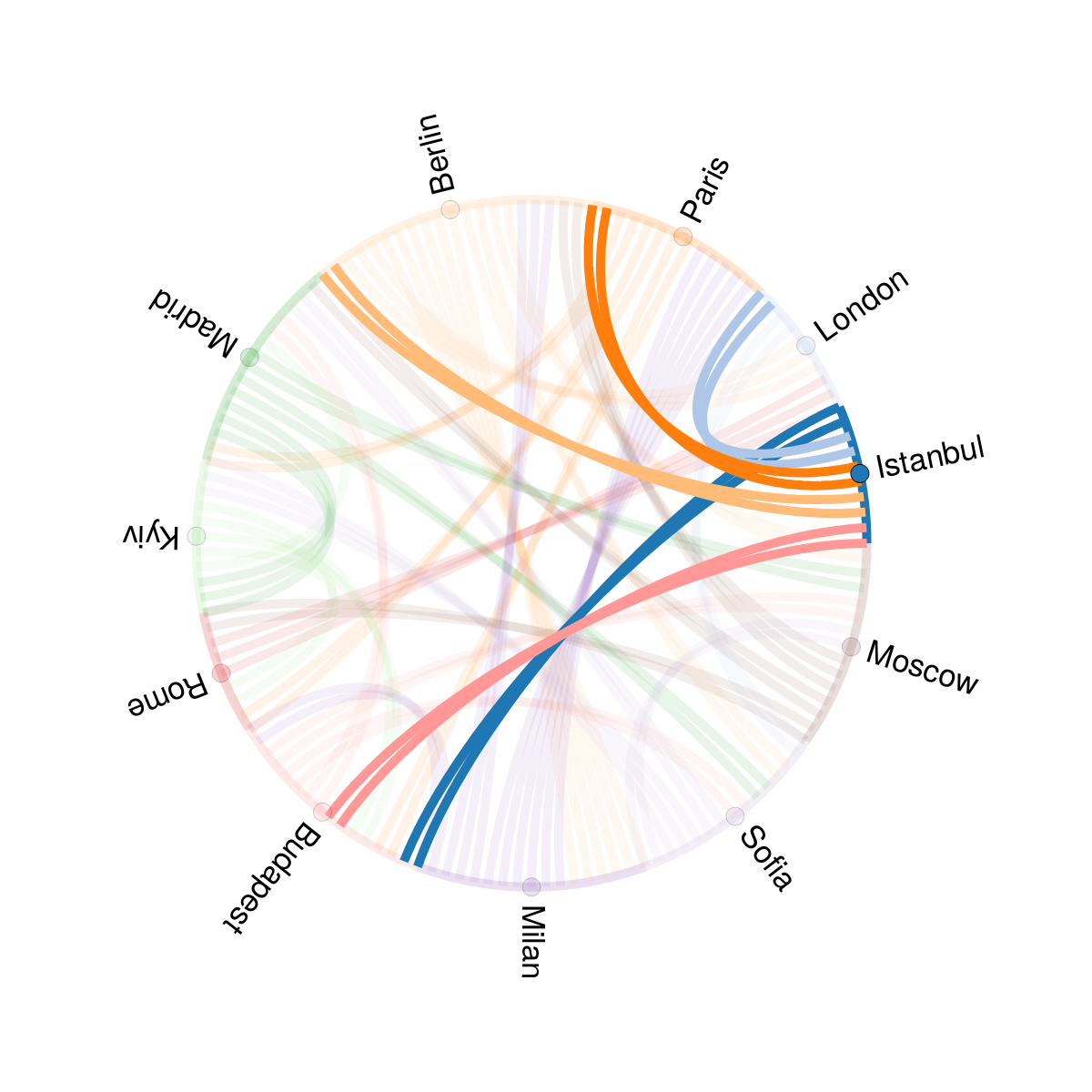}
    \caption{Istanbul}\label{fig:infl_cities_style_c}
\end{subfigure}\\
\begin{subfigure}{.32\textwidth}
    \centering
    \includegraphics[trim={3cm 4cm 3.5cm 3.5cm} ,clip,width=0.98\linewidth]{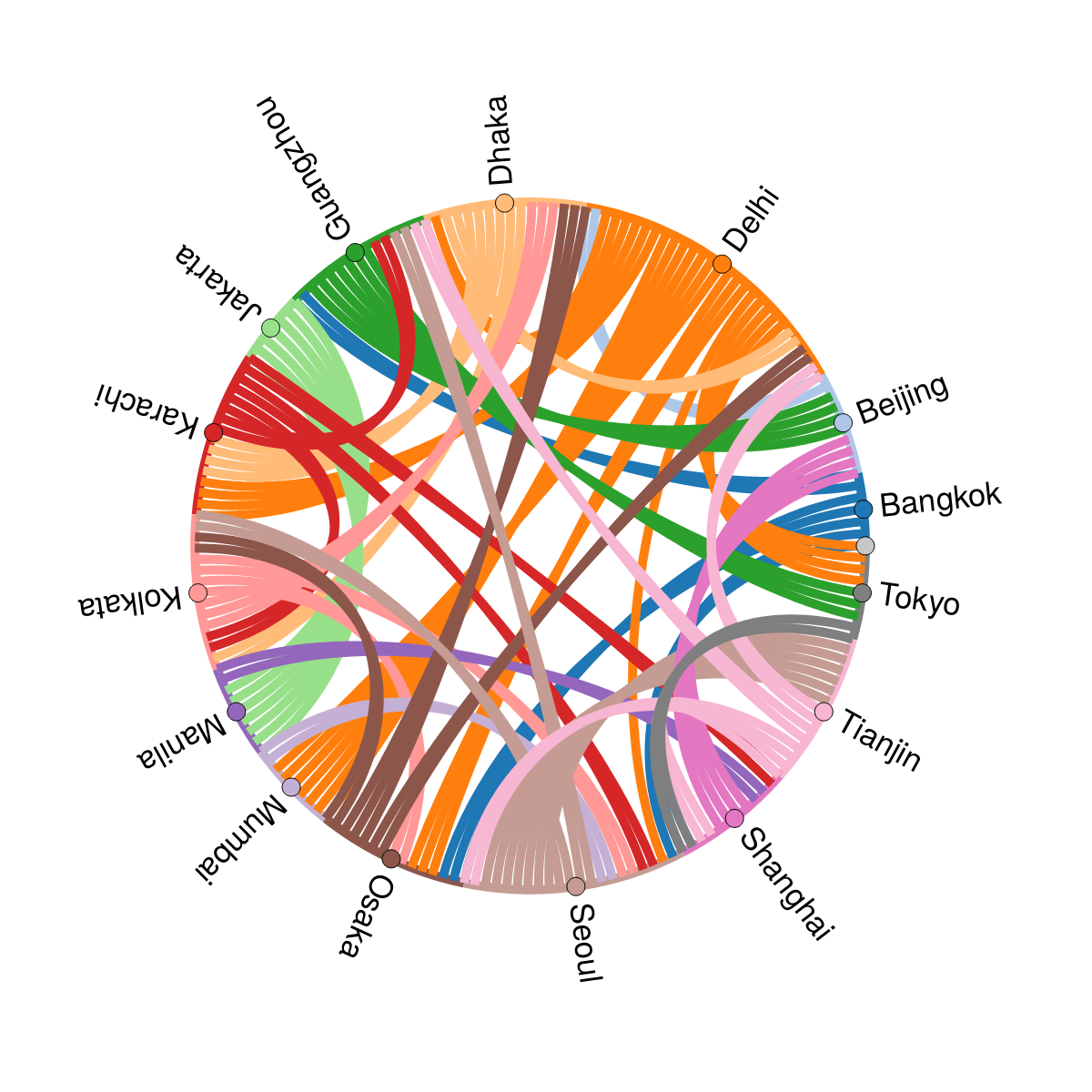}
    \caption{Asian Cities}\label{fig:infl_cities_style_d}
\end{subfigure}
\begin{subfigure}{.32\textwidth}
    \centering
    \includegraphics[trim={3cm 4cm 3.5cm 3.5cm} ,clip,width=0.98\linewidth]{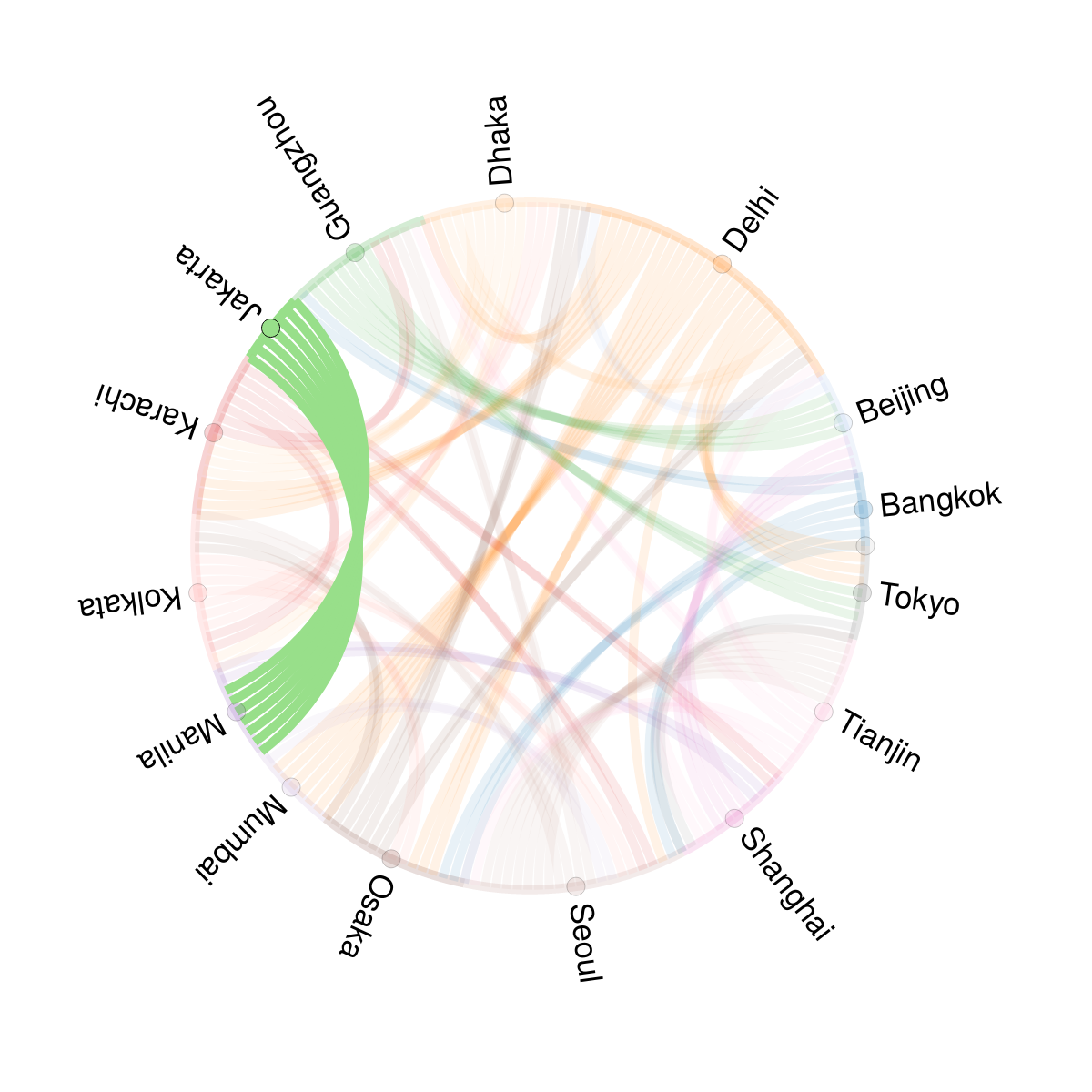}
    \caption{Jakarta}\label{fig:infl_cities_style_e}
\end{subfigure}
\begin{subfigure}{.32\textwidth}
    \centering
    \includegraphics[trim={3cm 4cm 3.5cm 3.5cm} ,clip,width=0.98\linewidth]{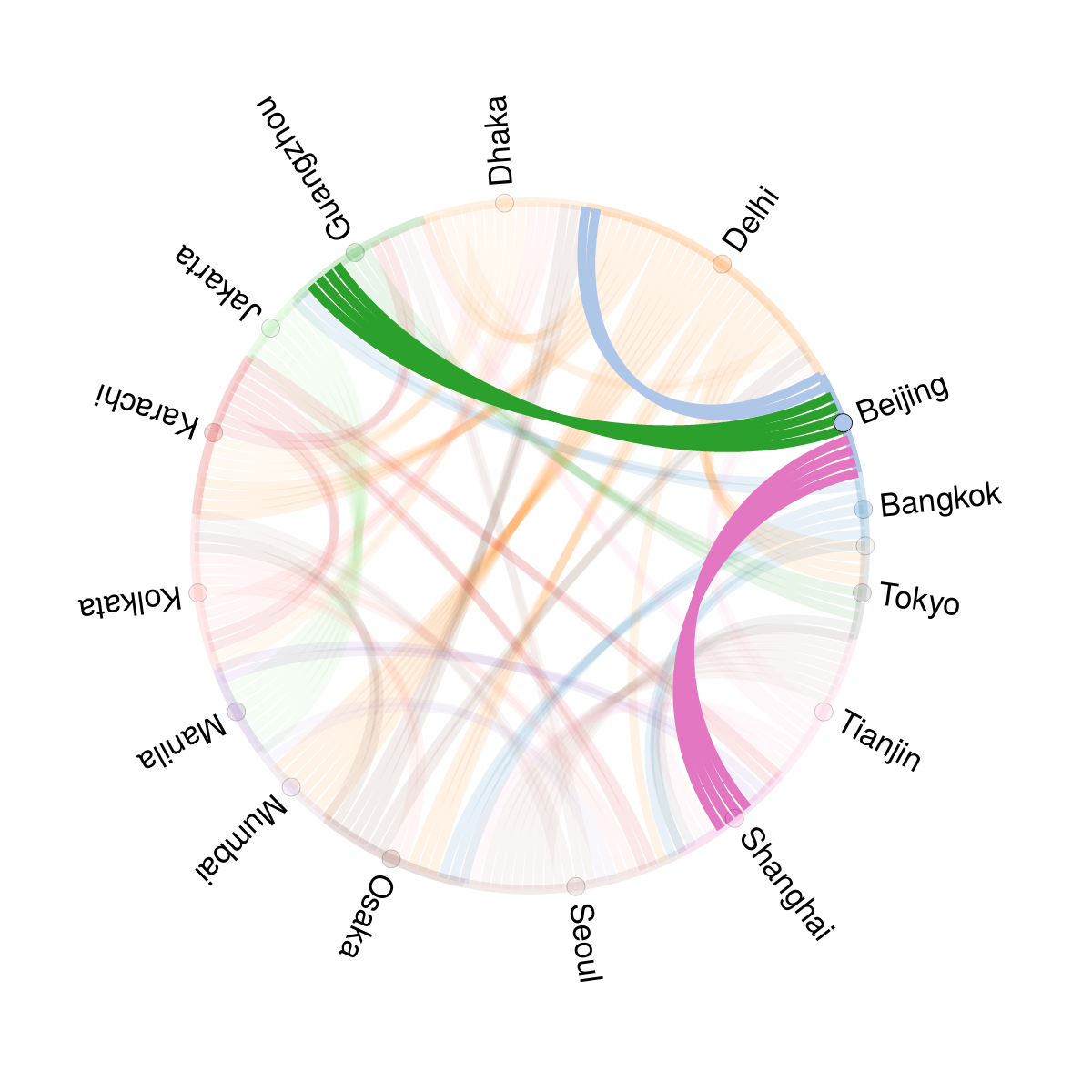}
    \caption{Beijing}\label{fig:infl_cities_style_f}
\end{subfigure}

\caption{Style influence relations discovered by our model among European (a) and Asian (d) cities.
The number of chords coming out of a node (\ie a city) is relative to the influence weight of that city on the receiver.
Chords are colored according to the source node color, \ie the influencer.
Our model discovers various types of influence relations from multi-city (\eg Paris) and single-city (\eg Jakarta) influencers to cities that are mainly influence receivers (\eg Istanbul and Beijing).
}
\label{fig:infl_cities_style}
\vspace*{-0.1in}
\end{figure*}
 
\tblref{tbl:forecast_geostyle_ext} shows the results.  
The proposed model outperforms all the naive, per-city, and all-cities models.
This shows the value of discovering influence for the quantitative forecasting task.
Ablation studies (bottom segment of the table) show the impact of each component of our model.
We evaluate two versions of our model, one without our influence modeling from \secref{sec:app_influence}, which assumes a full interaction pattern among all cities, and a second version that also is not trained for coherent forecasts (\secref{sec:app_coherence}).
We see that both our influence modeling and coherence-based forecasts are important for accurate predictions.

We notice that the styles' popularity trajectories do have a seasonal component; seasonal models (like GeoModel~\cite{mall2019geostyle} and Seasonal) do well compared to non-seasonal ones (like AR and Drift), but still underperform our approach.
This ranking changes on the deseasonalized test, where models like FashionForward~\cite{Al-Halah2017} and AR outperform the seasonal ones.
Our model outperforms all competitors on both types of trajectories, which demonstrates the benefits of accounting for influence.  
Our model's improvements compared to the best ``non-influence'' per-city or all-cities competitors on both types of style trajectories are statistically significant based on a t-test with $p<0.05$.

\begin{figure}[t]
\centering
\begin{subfigure}{.49\textwidth}
\includegraphics[width=0.98\linewidth]{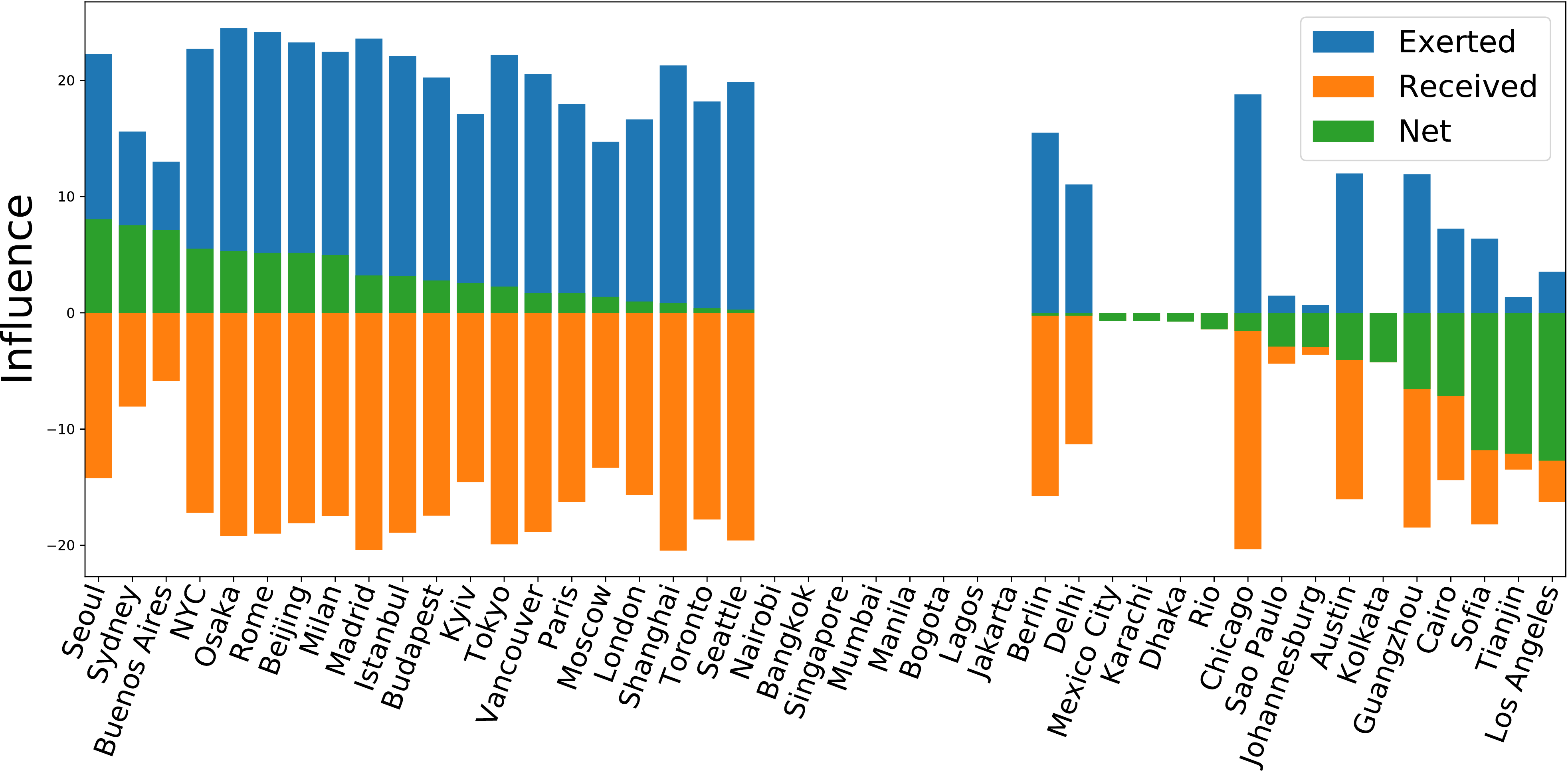}
\end{subfigure}
\begin{subfigure}{.49\textwidth}
\includegraphics[width=0.98\linewidth]{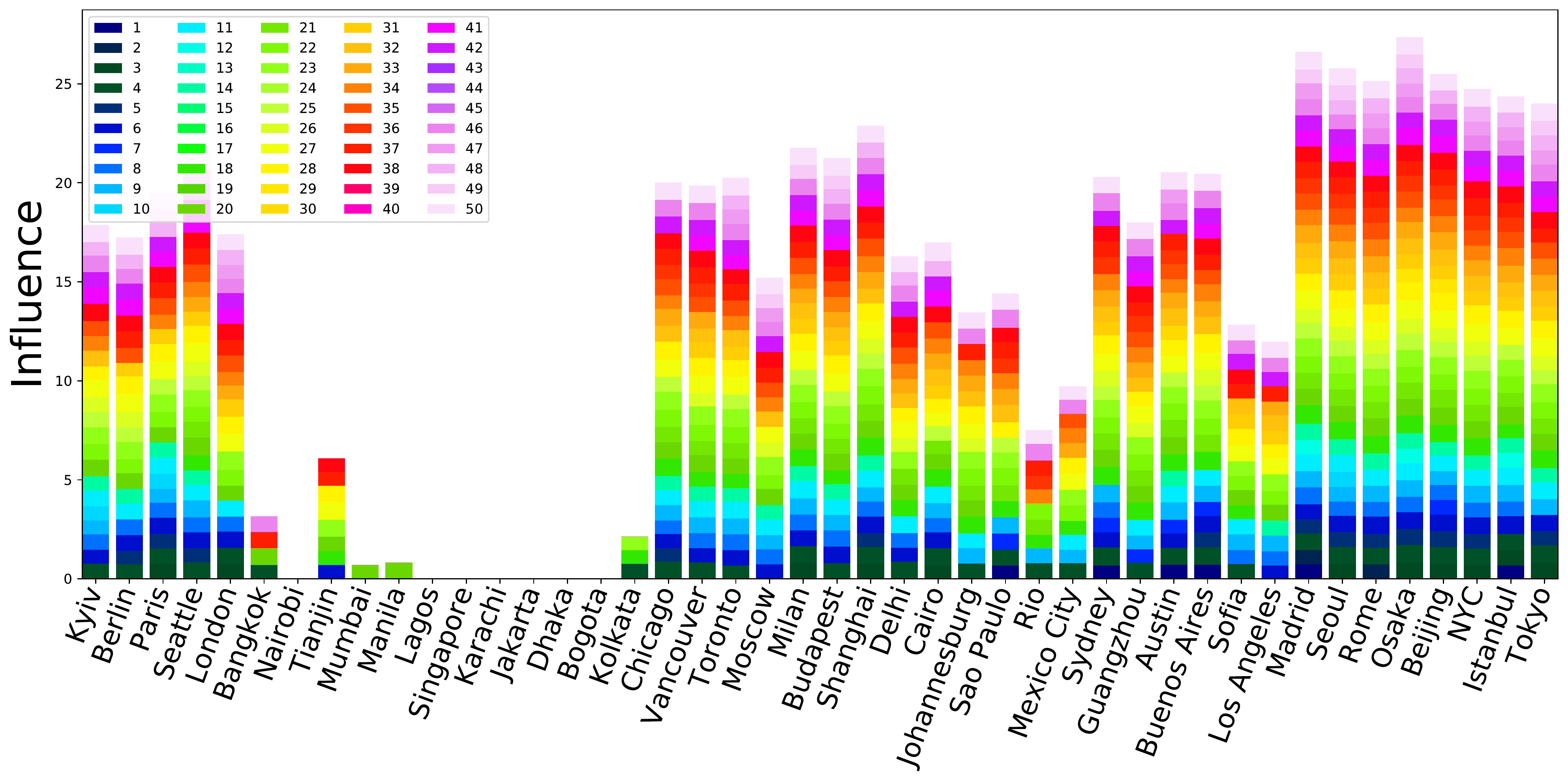}
\end{subfigure}
\vspace*{-0.05in}
\caption{
Top: Worldwide ranking of major cities according to their fashion influence on their peers.
The cities are sorted by their \emph{net} influence score (green).
The center cities with no bars indicate that they do not have influential relations with above average weight.
Bottom: the exerted influence score is split into influence score per individual style for each city (sorted by style influence similarity).
}
\label{fig:world_infl_rank_cities_style}
\end{figure}

\subsection{Influence Relations}\label{sec:eval_relations}

The results thus far confirm that our method's discovered influence patterns are meaningful, as seen by their positive quantitative impact on forecasting accuracy.  Next, we analyze them qualitatively to understand more about what was learned.
We consider influence interactions along two axes: 1) a local one that looks at pairwise influence relations among the cities; and 2) a global one which examines how cities influence the world's fashion trends.

\paragraph{1) City $\rightarrow$ City influence}
For each visual style, our model estimates the influence relation between cities and at which temporal lag, yielding a tensor $\mathbf{B}\in\mathbb{R}^{|C|\times|C|\times|S|}$ such that $B^{k}_{ij}$ is the influence lag of $C^i$ on $C^j$ for style $S^k$.
By averaging these relations across all visual styles, we get an estimate of the overall influence relation between all cities weighted by the temporal length, \ie long term influencers are given more weight than instantaneous ones. 
We visualize the influence relations using a directed graph where each node represents a city, and we create a weighted edge from city $C^i$ to $C^j$ if $C^i$ is found to be influencing $C^j$.

\figref{fig:infl_cities_style} shows an example of the influence pattern for fashion styles discovered by our model among major European (top left) and Asian (bottom left) cities, where the number of connections between two cities is relative to the weight of the influence relation.
Our model discovers interesting patterns. For example, there are a few fashion hubs like \emph{Paris} and \emph{Berlin} which exert influence on multiple cities while at the same time being influenced by few (one or two) cities. \emph{Paris} influences $4$ cities in Europe while being influenced by \emph{Milan} only (\figref{fig:infl_cities_style_b}).
Cities like \emph{Jakarta} have a one-to-one influence relation with \emph{Manila} (\figref{fig:infl_cities_style_e}).
On the other end of the spectrum, we find cities like \emph{Istanbul} (\figref{fig:infl_cities_style_c}) and \emph{Beijing} (\figref{fig:infl_cities_style_f}) that mainly receive influence from multiple sources while influencing few.

Additionally, we rank all cities according to their accumulated influence power on their peers.  
That is, we assign an influence score for each city according to the sum of weighted influence relations \emph{exerted} by that city on the rest.
Similarly, we also calculate the sum of \emph{received} influence as well as the difference in both as the \emph{net} influence score.
\figref{fig:world_infl_rank_cities_style} (top) shows these three influence scores for all cities across the world, sorted by the \emph{net} score.
The ranking reveals that some cities, like \emph{London} and \emph{Seattle}, act like focal points for fashion styles, \ie they receive and exert a high volume of influence simultaneously.
Others, like \emph{Seoul} and \emph{Osaka}, have a high net influence, which could indicate having some unique fashion styles not influenced by external players.
Furthermore, breaking down the exerted influence score for each city to per-style influence scores, we see in \figref{fig:world_infl_rank_cities_style} (bottom) that we can identify and group influencers into ``teams" based on their common set of styles where they exert their influence.
For example, \emph{Chicago}, \emph{Vancouver}, and \emph{Toronto} constitute a team since they seem to be influencing similar sets of visual styles.

\begin{figure}[t]
\centering
    \includegraphics[width=0.90\linewidth]{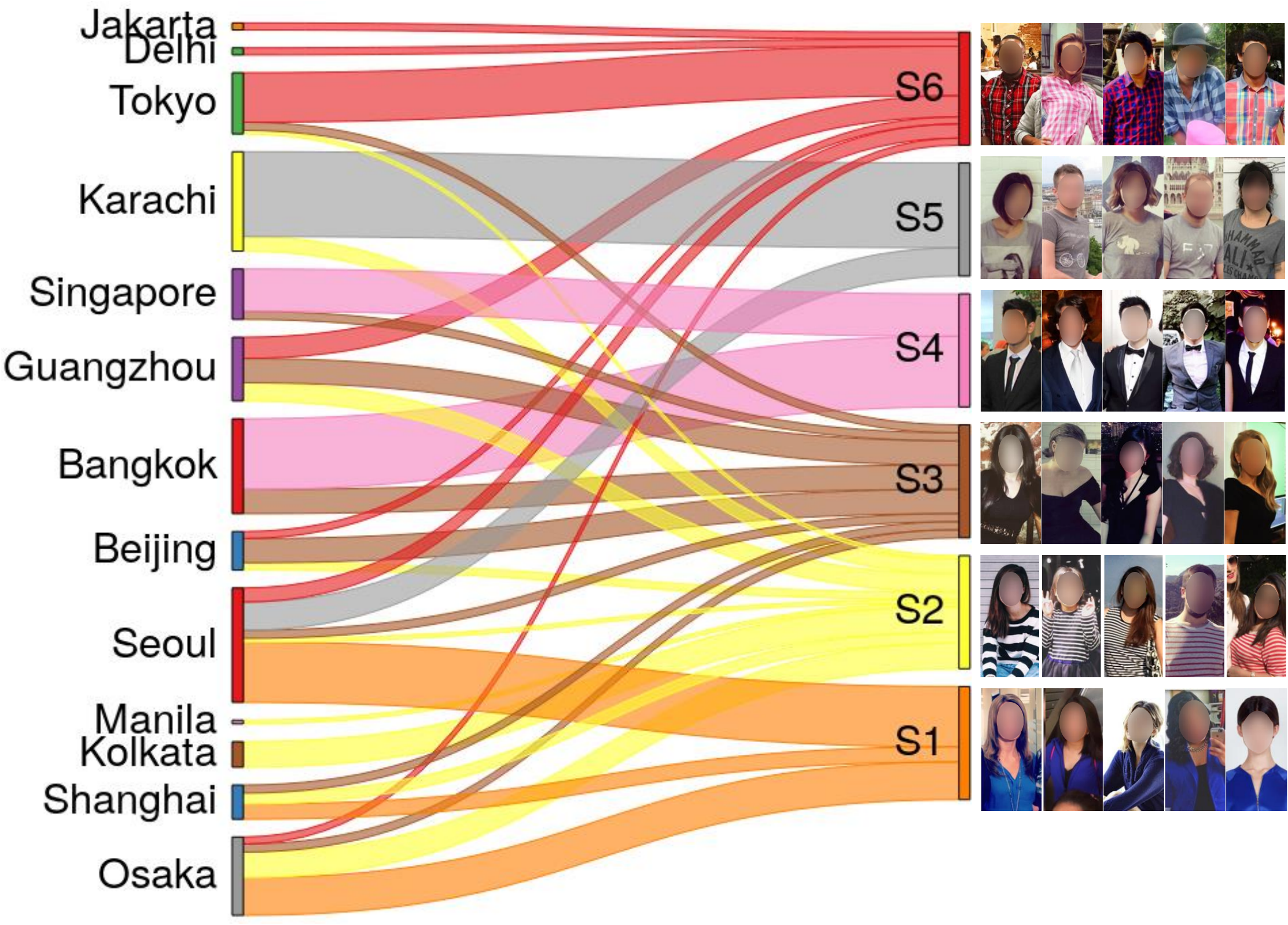}
    \vspace*{-0.1in}
\caption{Discovered influence by our model of Asian cities on global trends of $6$ fashion styles.
The width of the connection is relative to the influence weight of that city in relation to other influencer of the same style.
}
\label{fig:infl_asia_cities_global_styles}
\end{figure}
 
\vspace*{-0.05in}
\paragraph{2) City $\rightarrow$ World influence}
Alternatively, we can analyze the influence relation between a city and the world trend for a specific style.
This helps us better understand who are the main influencers on the world stage for each of the styles.
We capture this relation by modeling the interaction of a city's popularity trajectory on the global one (\ie the observed trend of the collective popularity of the same style across the world).

\figref{fig:infl_asia_cities_global_styles} shows a set of Asian cities and their influence on six global fashion styles.
We see that for some of the fashion styles, like $S_4$ and $S_5$, a couple of cities maintain a monopoly of influence on them, whereas others, like $S_3$ and $S_2$, are influenced almost uniformly by multiple cities.
Our influence model also reveals the influence strength (measured by the temporal lag) of these cities relative to their peers at the world stage.
See for example the strong influence of \emph{Seoul} and \emph{Bangkok} compared to the delicate one of \emph{Manila} and \emph{Jakarta}, as represented by the width displayed for their respective influence relation to the global trends.

\subsection{Influence Correlations}\label{sec:correlations}

Next we analyze the correlations of these relations discovered by our model with known real-world properties of the cities. 
We stress that the trends visible in the photos \emph{are} exactly what our model measures; there is no separate ``ground truth" against which to score the influence measurements.  Correlating against other properties simply helps unpack what the trends do or do not relate to.
We collect information about the annual gross domestic product (GDP), the geolocation, the population size, and the yearly average temperature for each of the cities.
We calculate the correlations of these properties with the influence information discovered by our model at two levels:
1) influence world ranking (\ie does a high influence rank correlate with the population size of the city? do cities with warm weather have a high influence score?), and 2) relation direction (\ie does influence flow from high to low GDP cities? do cities influence those that are geographically close to them?).

\begin{table}
\setlength{\tabcolsep}{12pt}
\center
\scalebox{0.9}{
\begin{tabular}{l r r}
\toprule
                    & \multicolumn{2}{c}{Fashion Influence} \\
    Meta Info.      &  Direction    &   World Rank \\
\midrule
    GDP             &   0.037       &   0.373   \\
    Temperature     &   - 0.319     &   - 0.616   \\
    Latitude        &   - 0.348     &   0.596   \\
    Population      &   0.038       &   - 0.193   \\
    Distance        &   - 0.165     &   n/a     \\
    Num. Samples    &   - 0.148     &   0.086   \\
\bottomrule
\end{tabular}
}
\caption{
Correlations of the discovered influence patterns 
with meta information about the cities.
} 
\label{tbl:metacorr}
\end{table}
 
\paragraph{World rank}
In \tblref{tbl:metacorr} ($2^{nd}$ column) we correlate the discovered influence ranking of all the cities with the ranking derived from each of the meta properties using the Spearman coefficient.
The correlation of these meta properties with the overall ranking of the cities uncovers some curious cases.
There is an above average correlation between the city influence rank and its latitude; many of the influential fashion cities are on the northern hemisphere.
We see a weaker but positive correlation with GDP, \ie a higher GDP could be a faint indicator of a higher fashion influence.
Finally, we observe a negative correlation with average temperature; influential fashion cities are often colder.

\paragraph{Relation direction}
\tblref{tbl:metacorr} ($1^{st}$ column) shows the correlation of the influence directions discovered by our model with differences in each of the meta properties between the influencer and the influenced city. 
Specifically, for each city and meta property $M_i$ (\eg GDP), first we measure the differences between that city and the rest in regards to $M_i$, then we correlate these differences with the influence exerted by that city.
Interestingly, none  show high correlation with fashion influence directions.
The relation type cannot be reliably estimated based on the differences in GDP (\eg high GDP cities do not always influence lower GDP ones), population (\eg cities with high population do not necessarily influence others with lower population or vice versa), nor  distance (\eg influence does not correlate well with how far one city is from its influencer).
A weak and negative correlation is found with temperature and latitude differences, showing that cities with similar temperature or at similar latitudes tend to influence each other slightly more.
These results suggest our model discovers complex fashion influence relations that are hard to infer from generic properties of the constituent players.

As a sanity check, we also explore the correlation of the number of image samples collected from each city in the dataset with the two types of influence information (\tblref{tbl:metacorr} last row).
We find that there is no strong correlation between the learned influences and the number of images available in the data for each city (\ie influential cities are not those with a higher number of samples in the dataset).

\subsection{Influence Dynamics} 

\begin{figure}[t]
\centering
    \includegraphics[width=0.65\linewidth]{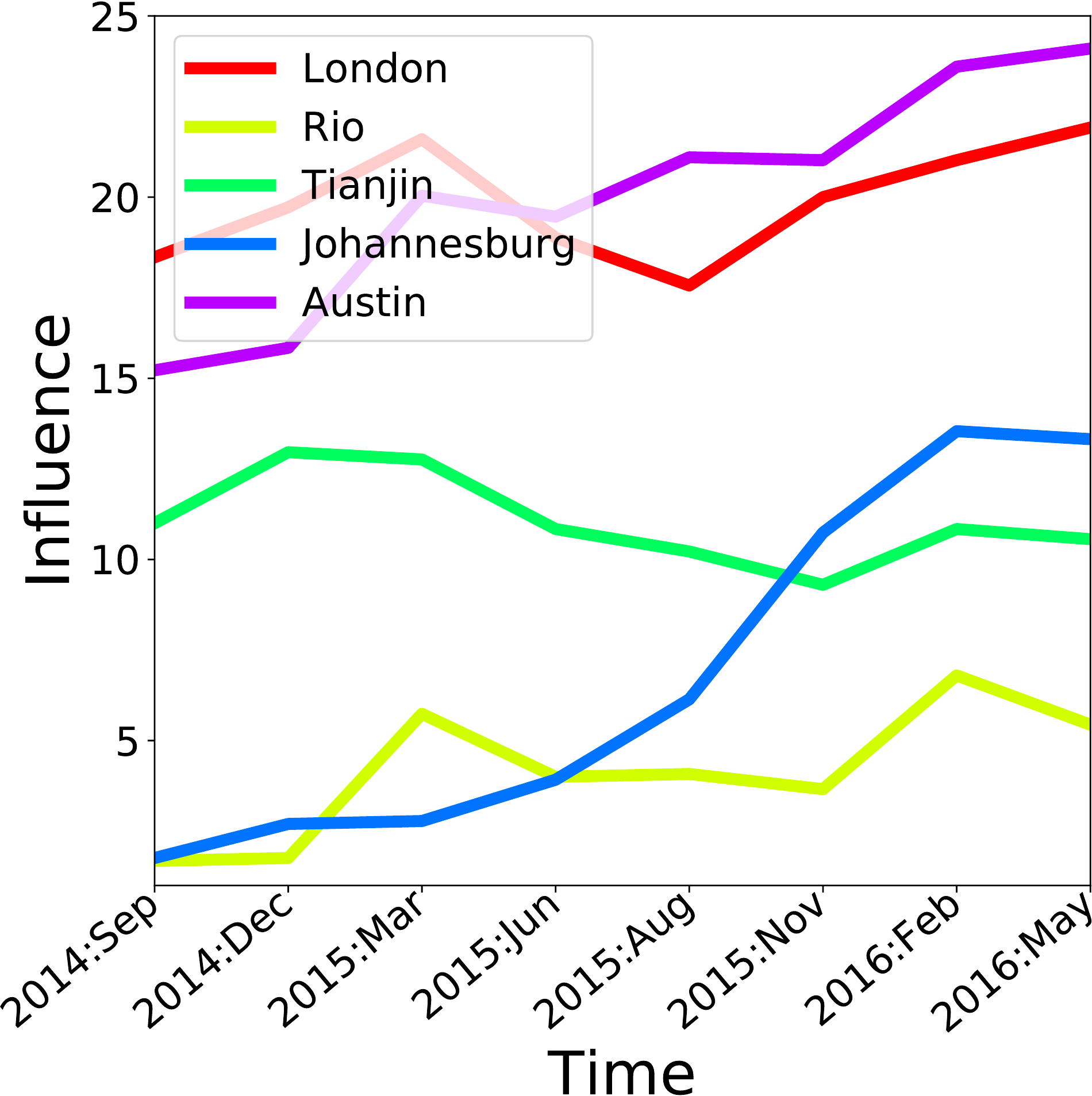} 
    \vspace*{-0.1in}
\caption{Dynamics analysis of exerted fashion influence at multiple time steps (with a 3 month interval) reveals the cities' temporal changes in influence strength. 
}
\label{fig:world_infl_dynamics}
\end{figure}
 
Finally, we study the changes in the influence rank of the cities through time.
We carry out our influence modeling based on the style trajectories of the various cities as before, but at multiple sequential time steps.
Then we collect the overall influence score of each city at each step. 

\figref{fig:world_infl_dynamics} shows the change in the influence score for a subset of $5$ cities spanning different continents.
We notice that cities show various dynamic behaviors across time.
While some cities like \emph{London} and \emph{Rio} maintain a steady influence score through time (at different levels), others like \emph{Austin} and \emph{Johannesburg} demonstrate a positive trend and are gaining more influence in the fashion domain over time but at varying speeds.
Other cities like \emph{Tianjin} exhibit a mild decline in their fashion influence.

\vspace*{-0.05in}
\section{Conclusion}

We introduced a model to quantify influence of visual fashion trends, capturing the spatio-temporal propagation of styles around the world.
Our approach integrates both influence relations and a coherence regularizer to predict future style popularity conditioned on place.
Both our forecasting results and our analysis of the learned influences suggest potential applications in social science, where computer vision can unlock trends that are otherwise hard to capture.

\vspace*{0.05in}
\noindent\textbf{Acknowledgements:}  We thank Utkarsh Mall for helpful input on the GeoStyle data.  UT Austin is supported in part by NSF IIS-1514118.

{\small
\bibliographystyle{ieee_fullname}
\bibliography{mybib}
}
\vspace{1.5cm}
{\LARGE\noindent Appendix}
\appendix
\vspace{0.3cm}

\begin{figure}[b]
\centering
    \includegraphics[width=0.98\linewidth]{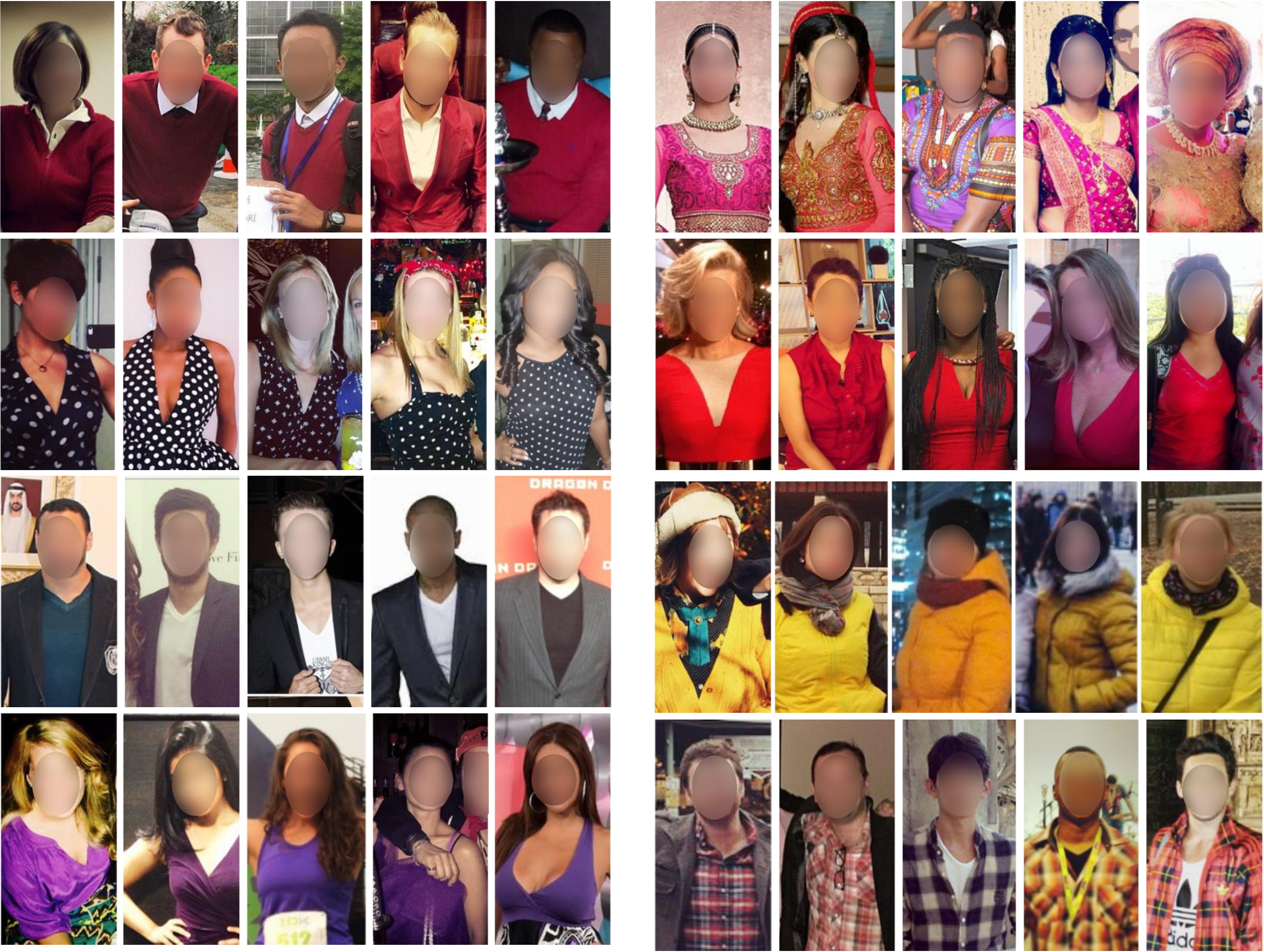}
\caption{
Examples of the learned fashion styles from a large-scale dataset of Instagram and Flickr images.
}
\label{fig:styles_examples}
\end{figure}
 \section{Fashion styles}
\figref{fig:styles_examples} shows more examples of the fashion styles learned by our style model from the GeoStyle dataset; a large-scale dataset of Instagram and Flickr images collected from $44$ major cities around the world.
A style is a combination of certain attributes describing materials, colors, cut, and other factors, such as \emph{V-neck, red, formal dress}.
Some of the learned fashion styles may reflect a season related type of garments (\eg the yellow jacket and scarf style) or a local traditional or cultural clothing (\eg \figref{fig:styles_examples} upper left row); however, many of the learned styles are common across different countries and cultures.

\section{Influence Relations}

\begin{figure}[hb]
\centering
\begin{subfigure}{.4\textwidth}
    \centering
    \includegraphics[width=0.98\linewidth]{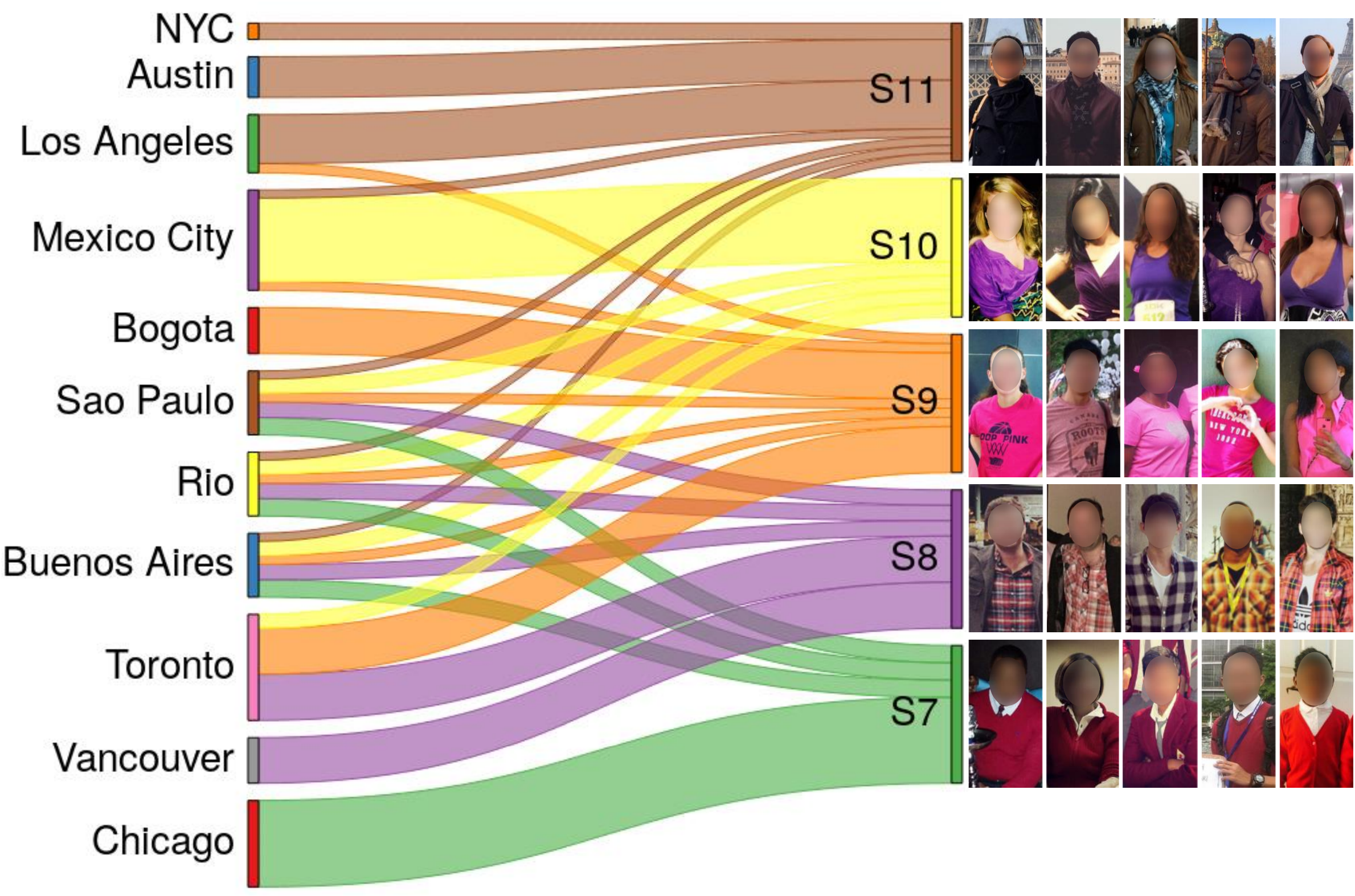}
    \caption{American Cities}\label{fig:infl_america_cities_global_styles}
\end{subfigure}\\
\begin{subfigure}{.4\textwidth}
    \centering
    \includegraphics[width=0.94\linewidth]{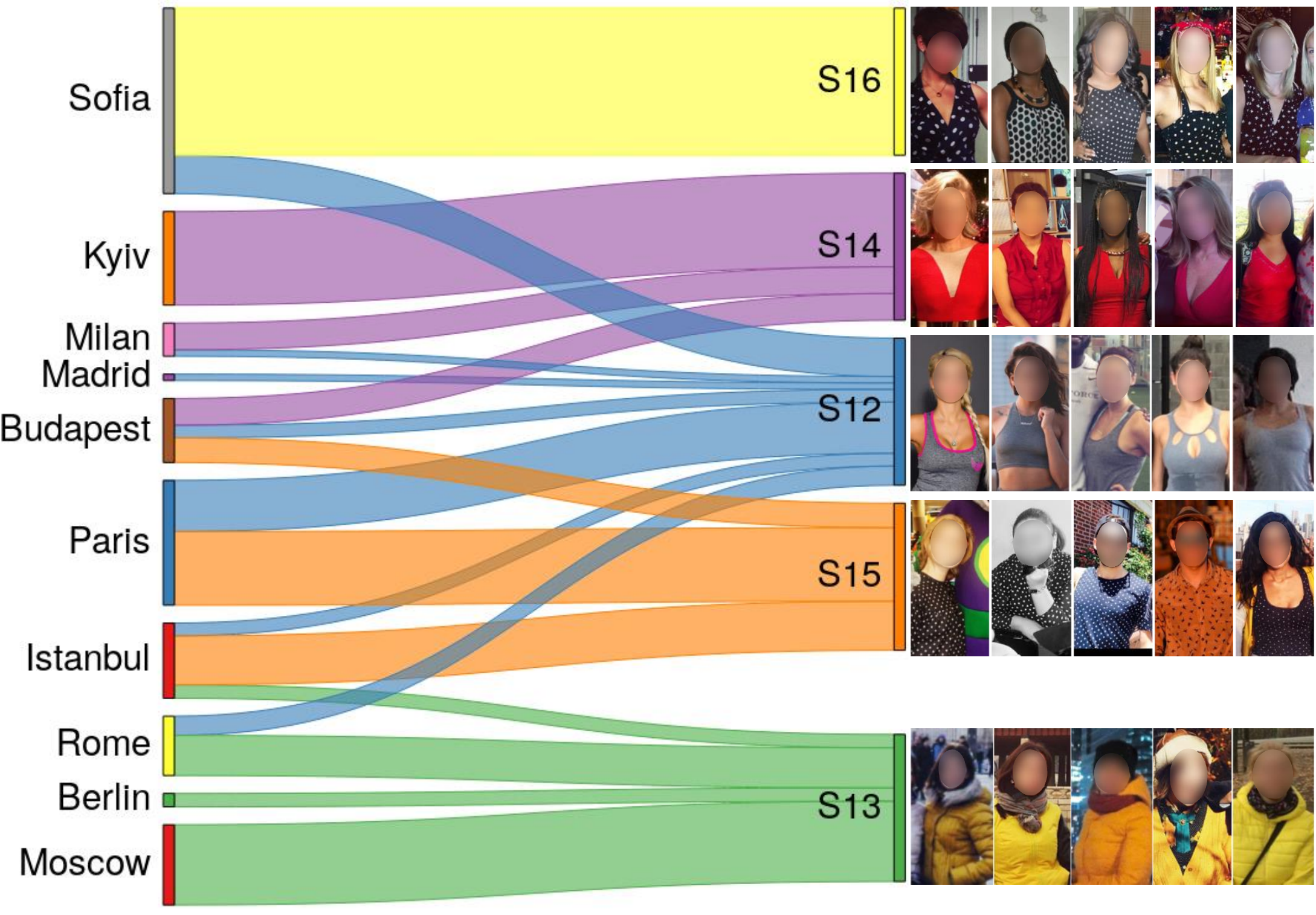}
    \caption{European Cities}\label{fig:infl_eu_cities_global_styles}
\end{subfigure}
\caption{
    Our model infers the fashion influence exerted by different cities on the global style trends, as seen here for (a) American and (b) European cities.
}
\label{fig:infl_america_eu_cities_global_styles}
\end{figure}
 \paragraph{City $\rightarrow$ World influence}
\figref{fig:infl_america_eu_cities_global_styles} shows the influence exerted by American (top) and European (bottom) cities on a set of learned fashion styles.
The figure reveals that some cities act as the main players and dominate the influence relation for the global trend of a fashion style, for example Chicago \& Style $S_7$ and Sofia \& Style $S_{16}$.
However, for few styles (like $S_9$ and $S_{12}$) the influence pattern is dispersed across multiple players.

\begin{figure*}[t]
\centering
\begin{subfigure}{.33\textwidth}
    \centering
    \includegraphics[trim={1.5cm 3cm 1.5cm 1.5cm} ,clip,width=1.\linewidth]{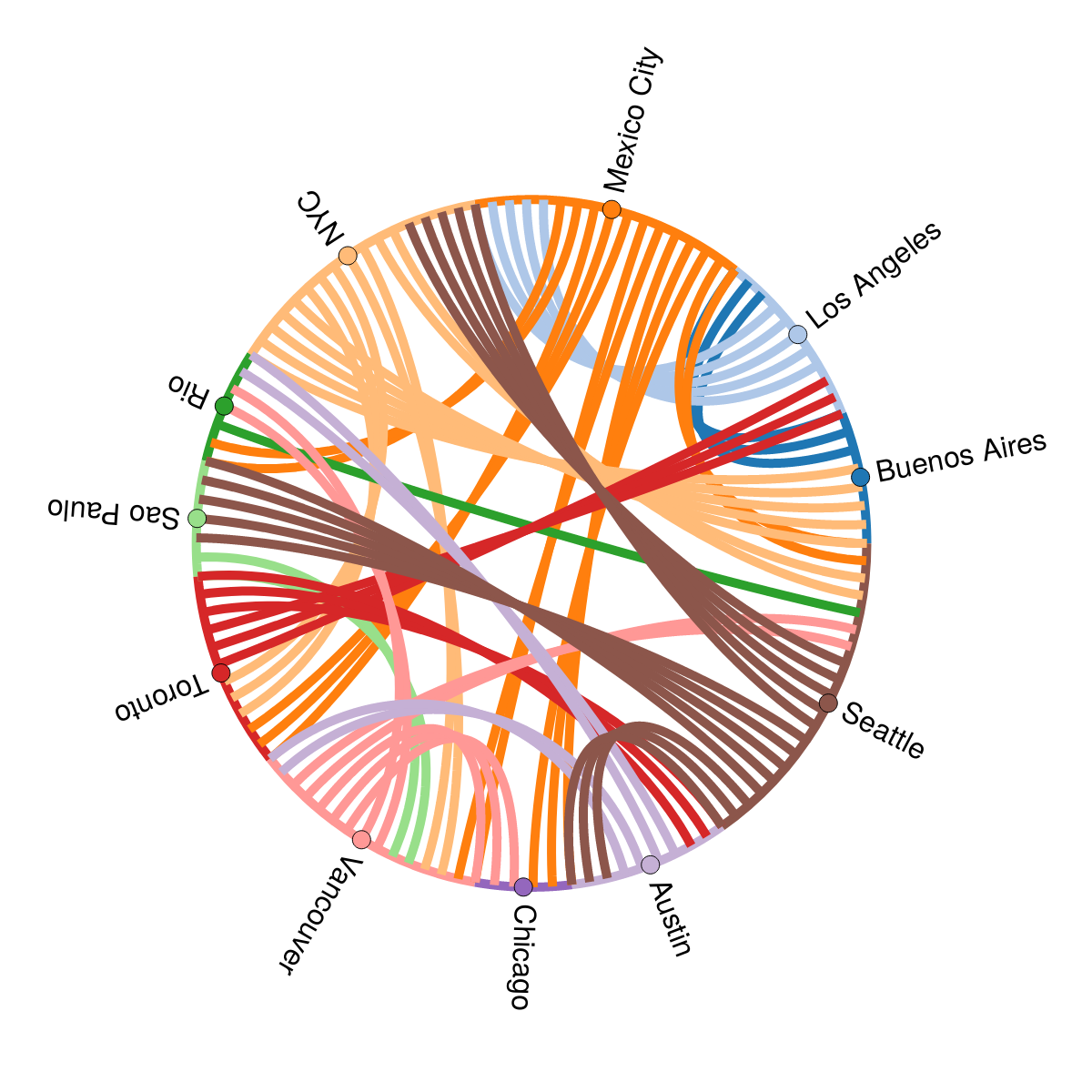}
    \caption{American Cities}
\end{subfigure}
\begin{subfigure}{.33\textwidth}
    \centering
    \includegraphics[trim={1.5cm 3cm 1.5cm 1.5cm} ,clip,width=1.\linewidth]{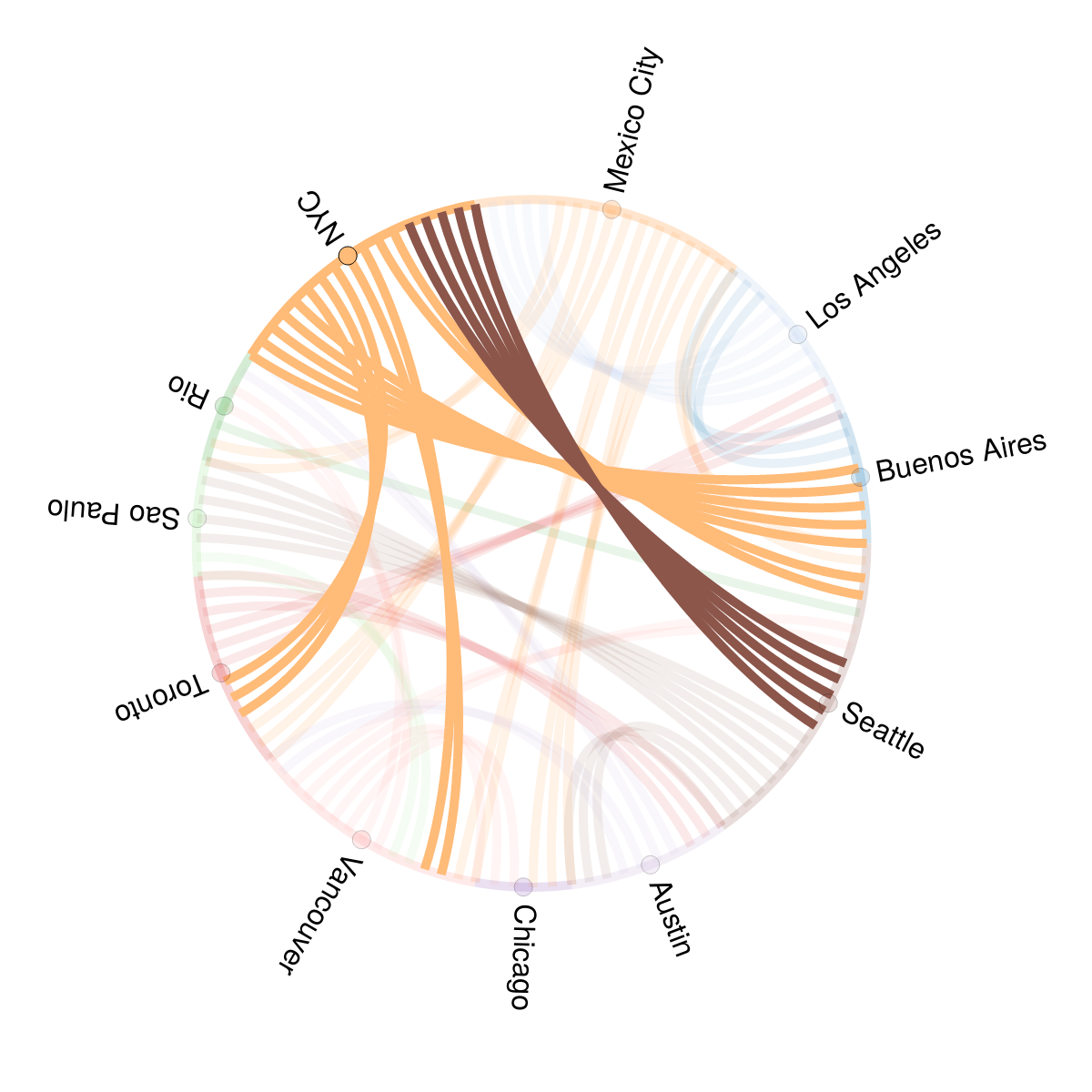}
    \caption{New York City}
\end{subfigure}
\begin{subfigure}{.33\textwidth}
    \centering
    \includegraphics[trim={1.5cm 3cm 1.5cm 1.5cm} ,clip,width=1.\linewidth]{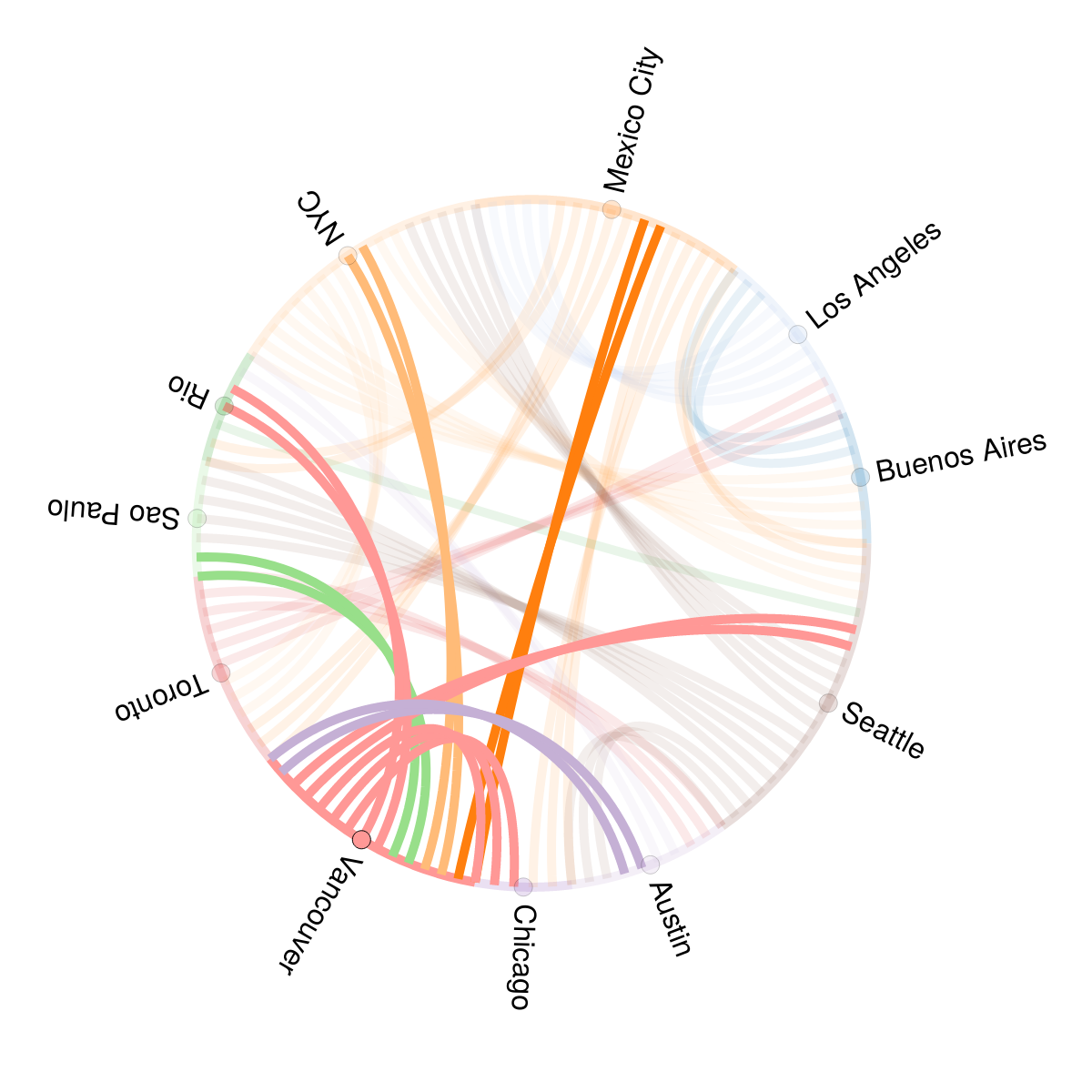}
    \caption{Vancouver}
\end{subfigure}\\
\begin{subfigure}{.33\textwidth}
    \centering
    \includegraphics[trim={1.5cm 3cm 1.5cm 1.5cm} ,clip,width=1.\linewidth]{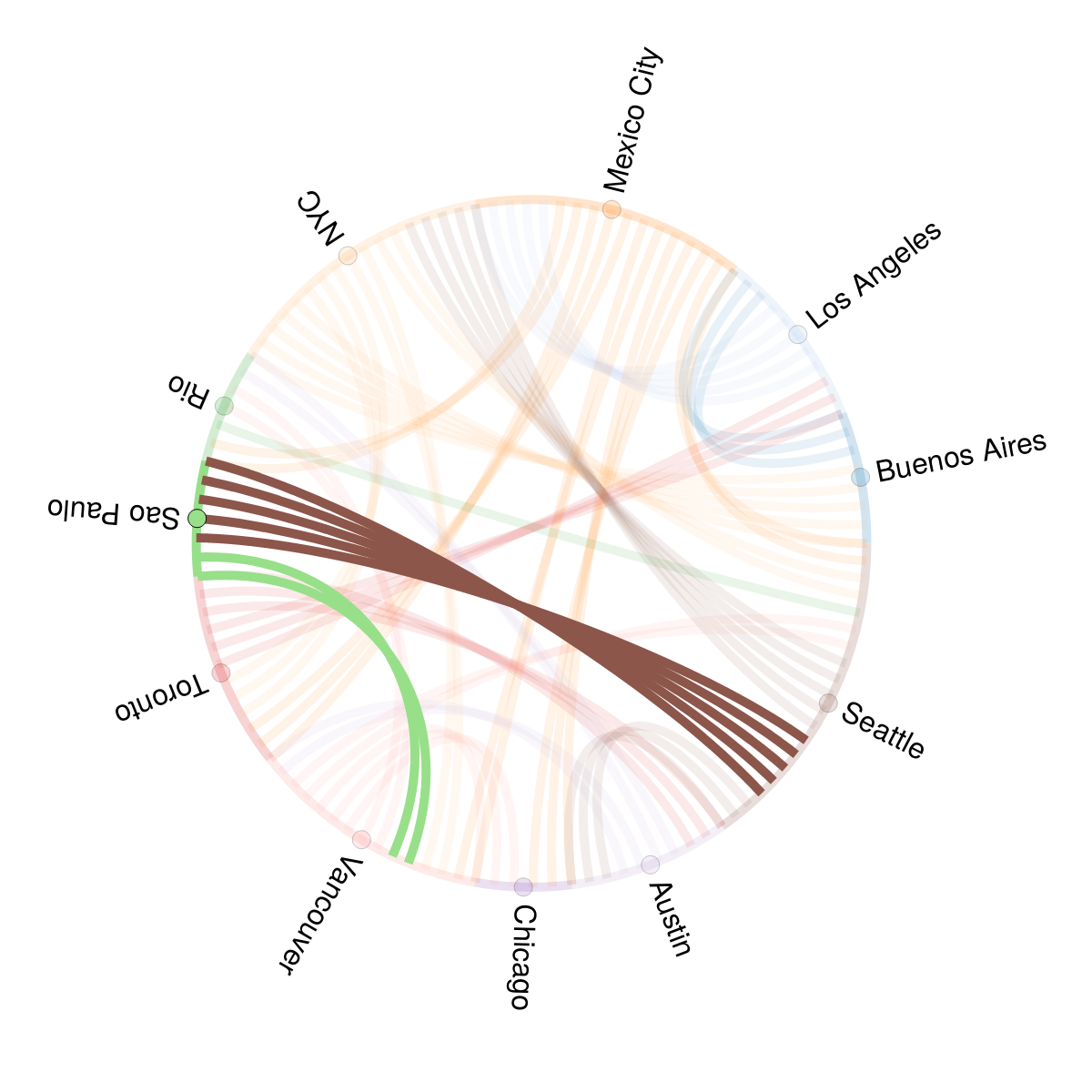}
    \caption{Sao Paulo}
\end{subfigure}
\begin{subfigure}{.33\textwidth}
    \centering
    \includegraphics[trim={1.5cm 3cm 1.5cm 1.5cm} ,clip,width=1.\linewidth]{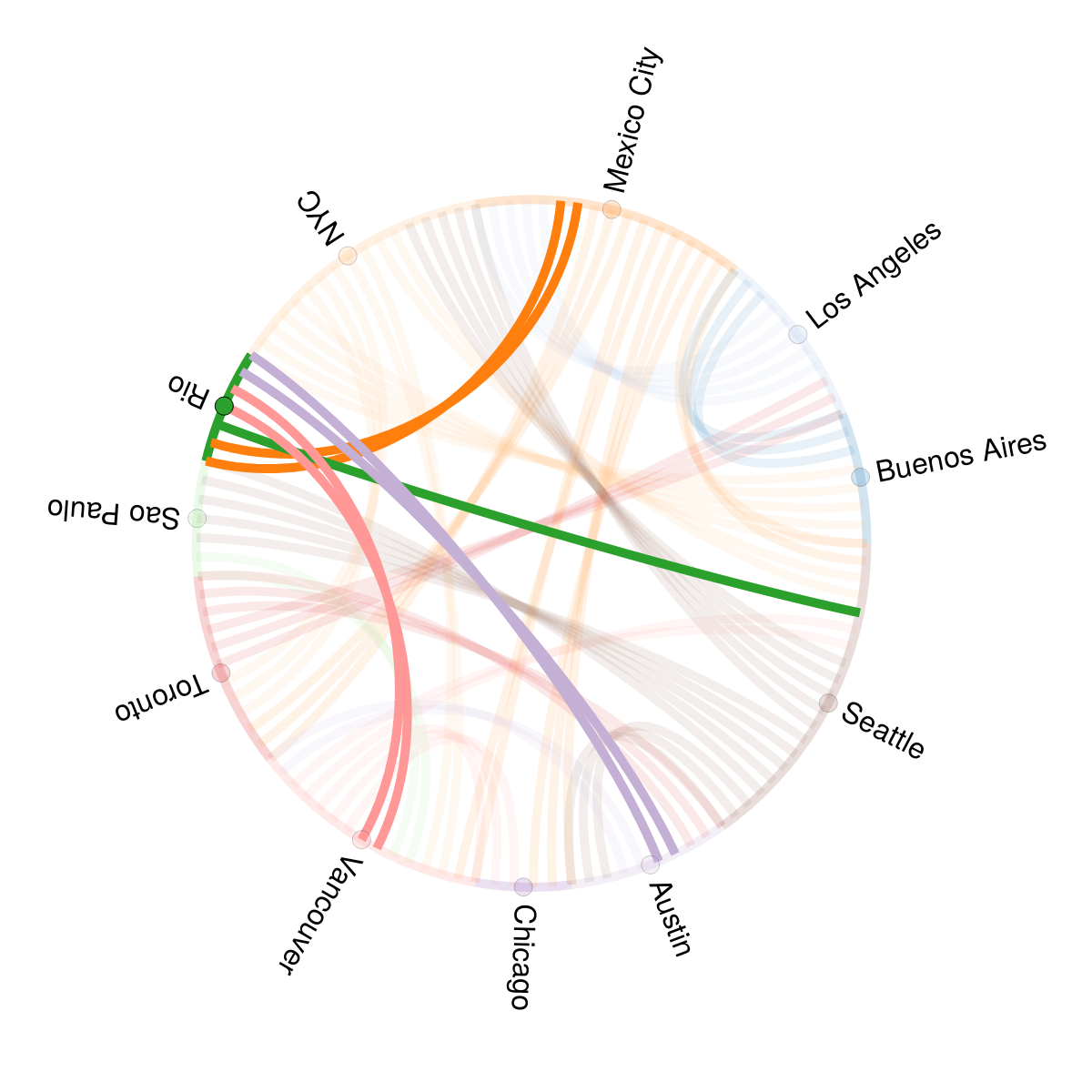}
    \caption{Rio de Janeiro}
\end{subfigure}
\begin{subfigure}{.33\textwidth}
    \centering
    \includegraphics[trim={1.5cm 3cm 1.5cm 1.5cm} ,clip,width=1.\linewidth]{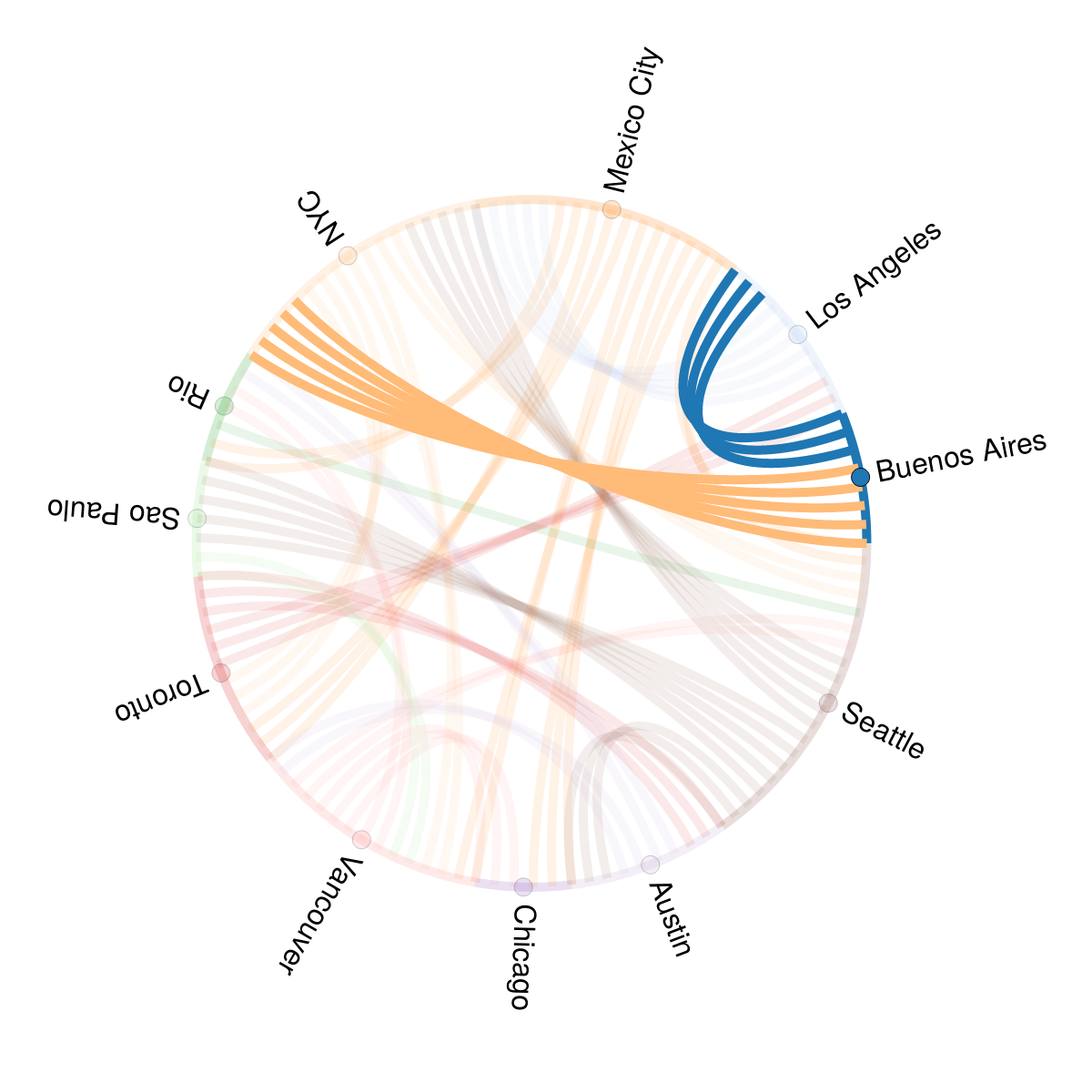}
    \caption{Buenos Aires}
\end{subfigure}

\caption{
Style influence relations discovered by our model among  cities in North and South America (top and bottom, respectively).
The number of chords coming out of a node (\ie a city) is relative to the influence weight of that city on the receiver.
Chords are colored according to the source node color, \ie the influencer.
Our model reveals interesting influence patterns, like the major fashion influencer (\eg New York City), the fashion hub (\eg Vancouver), and the single- or multi-source fashion receiver (\eg Buenos Aires and Rio de Janeiro, respectively).
}
\label{fig:infl_american_cities_style}
\end{figure*}
 \paragraph{City $\rightarrow$ City influence}
In \figref{fig:infl_american_cities_style}, we show the pairwise fashion influence relations discovered by our model on cities from the North and South American continents.
New York City is a major fashion influencer among the American cities and Vancouver exerts and receives fashion influence at a high volume indicating that it is an important fashion hub.
Other cities draw fashion influence from a main source (\eg Buenos Aires and Sao Paulo) or from a diverse set of influencer (\eg Rio de Janeiro).

\begin{figure*}[t]
\centering
\begin{subfigure}{.85\textwidth}
    \centering
    \includegraphics[width=1.\linewidth]{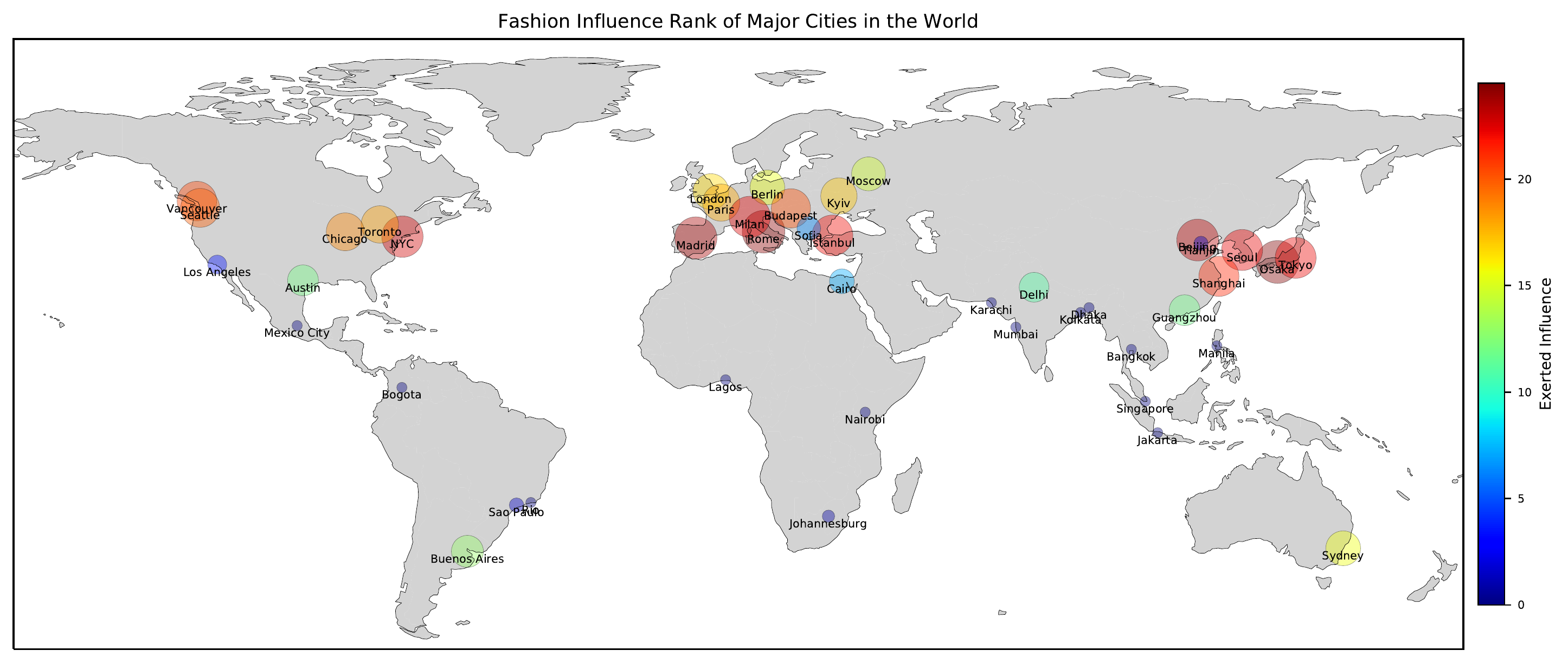}
    \caption{Major Cities}\label{fig:infl_map_city}
\end{subfigure}\\
\begin{subfigure}{.85\textwidth}
    \centering
    \includegraphics[width=1.\linewidth]{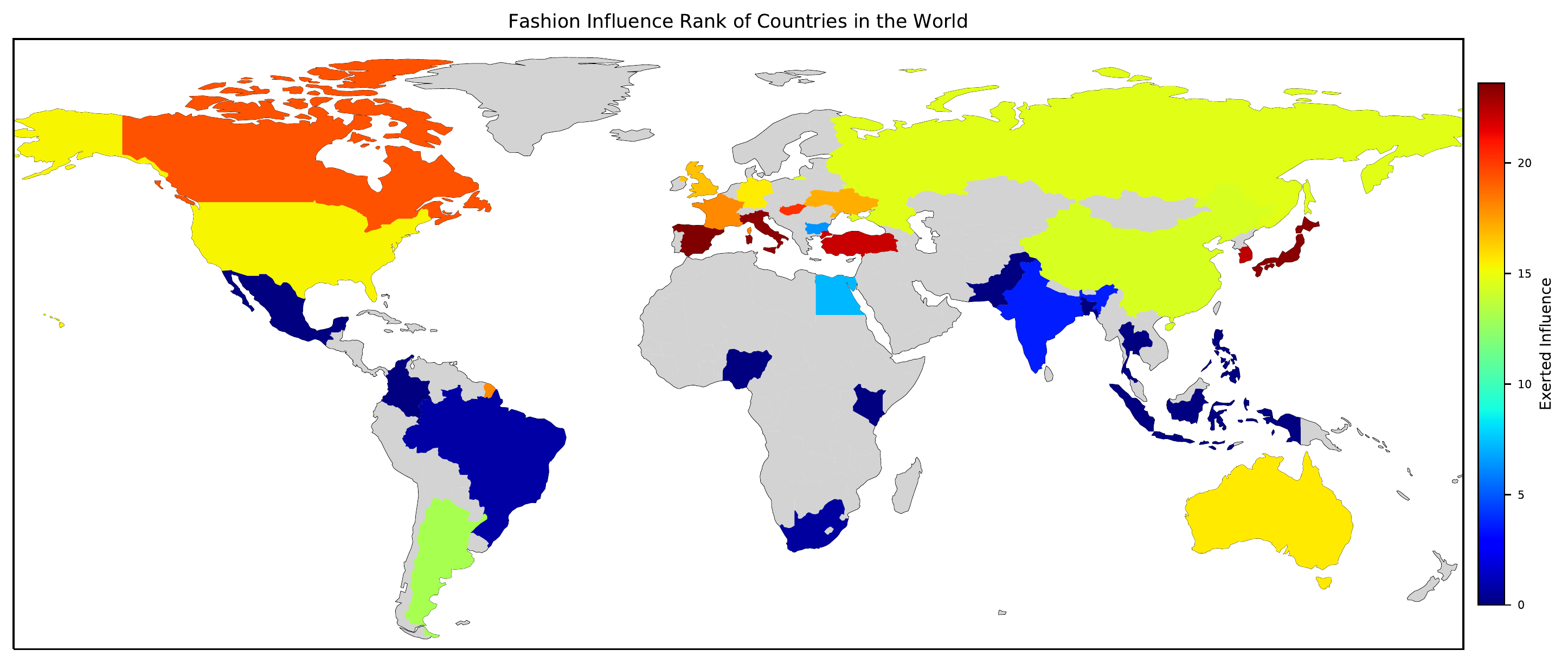}
    \caption{Countries}\label{fig:infl_map_country}
\end{subfigure}\\
\begin{subfigure}{.85\textwidth}
    \centering
    \includegraphics[width=1.\linewidth]{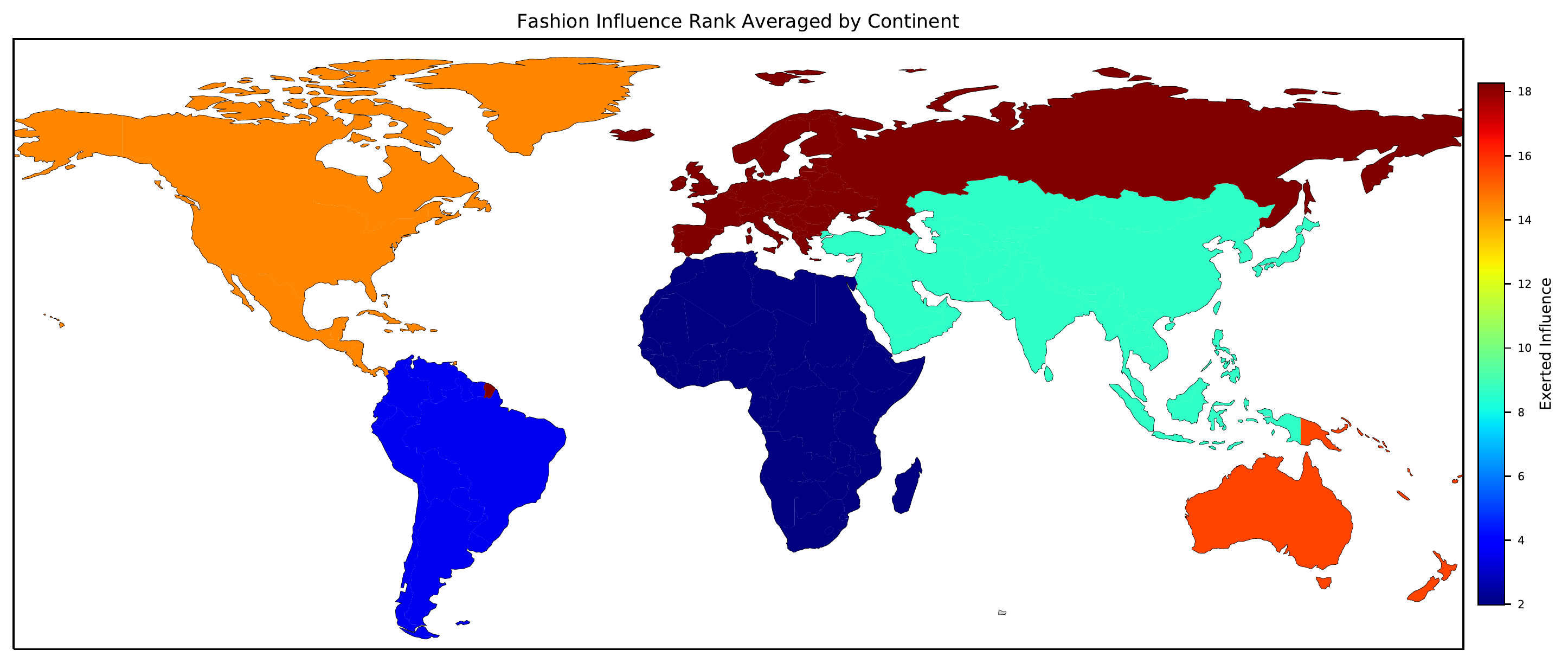}
    \caption{Continents}\label{fig:infl_map_continent}
\end{subfigure}

\caption{
    Fashion influence scores as inferred by our model from everyday images of people around the world at (a) the city, (b) the country, and (c) the continent level.
    Areas shaded with a gray color are ones without visual observations, consequently with no fashion influence estimates.
}
\label{fig:infl_worldmaps}
\vspace*{-0.15in}
\end{figure*}
 \section{A world view of fashion influence }
Finally, in \figref{fig:infl_worldmaps} we provide a global view of fashion influence on the world map.
\figref{fig:infl_map_city} reinforces our previous observation from the correlation analysis with cities meta information (Section 4.3 in the main paper).
Most of the influential cities lay in the northern hemisphere and specifically in its upper part, which also explains the positive correlation of the fashion influence rank with latitude and colder temperatures.
In \figref{fig:infl_map_country} and \figref{fig:infl_map_continent}, we show the influence score inferred by our model when aggregated by country and continent, respectively.
While at the country level we see that some Asian and American countries have high fashion influence scores (\eg Japan and Canada), at the continent level Europe still leads in terms of global fashion influence.

\section{Naive models}
In our forecast experiments, we consider five naive baselines that rely on basic statistical properties of the trajectory to produce a forecast:
\begin{itemize}[leftmargin=*,itemsep=2.5pt,labelsep=2pt,label=\textendash]
\vspace{-0.21cm}
\setlength{\parskip}{0pt}
\setlength{\parsep}{0pt}
    \item Gaussian: this model fits a Gaussian distribution based on the mean and standard deviation of the trajectory and forecasts by sampling from the distribution.\vspace*{-0.05in}
    \item Seasonal: this model forecasts the next step to be similar to the observed value one season before $y_{t+1} = y_{t-season}$.  We set a yearly season of 52 weeks.  \vspace*{-0.05in}
    \item Mean: it forecasts the next step to be equal to the mean observed values $y_{t+1} = \textrm{mean}(y_1,\dots,y_t)$.\vspace*{-0.05in}
    \item Last: it uses the value at the last temporal step to forecast the next $y_{t+1} = y_t$.\vspace*{-0.05in}
    \item Drift: it forecasts the next steps along the line that fits the first and last observations.
\end{itemize}

\end{document}